\documentclass[journal,comsoc]{IEEEtran}
\usepackage{epsfig}
\usepackage{epstopdf}
\usepackage{subfigure} 
\usepackage{amsmath}
\usepackage{graphicx}
\usepackage[T1]{fontenc}
\newcommand{\RN}[1]{%
	\textup{\uppercase\expandafter{\romannumeral#1}}%
}
\usepackage{cite}
\DeclareGraphicsExtensions{.pdf,.png,.jpg,.fig}
\usepackage[cmintegrals]{newtxmath}
\usepackage[linesnumbered,ruled]{algorithm2e}
\usepackage{multirow}
\begin{document}
	\title{Road Surface 3D Reconstruction Based on \\Dense Subpixel Disparity Map Estimation}

\author{Rui~Fan,~\IEEEmembership{Student Member,~IEEE,}  
	Xiao~Ai, Naim~Dahnoun 
        
\thanks{Rui Fan is with the Visual Information Group, the University of Bristol, BS8 1UB, UK. email: ranger.fan@bristol.ac.uk; ranger$\_$fan@outlook.com
	
	Xiao Ai is with the Quantum Technology Enterprise Centre, Nanoscience and Quantum Information Building, the University of Bristol, Bristol, BS8 1FD, UK. email: xiao.ai@bristol.ac.uk
	
	Naim Dahnoun is with the Department of Electrical and Electronic Engineering, Merchant Venturers Building, the University of Bristol, BS8 1UB, UK. email: naim.dahnoun@bristol.ac.uk}}

\markboth{IEEE TRANSACTIONS ON IMAGE PROCESSING}%
{Shell \MakeLowercase{\textit{et al.}}: Bare Demo of IEEEtran.cls for IEEE Communications Society Journals}
\maketitle

\begin{abstract}
Various 3D reconstruction methods have enabled civil engineers to detect damage on a road surface. To achieve the millimetre accuracy required for road condition assessment, a disparity map with subpixel resolution needs to be used. However, none of the existing stereo matching algorithms are specially suitable for the reconstruction of the road surface. Hence in this paper, we propose a novel dense subpixel disparity estimation algorithm with high computational efficiency and robustness. This is achieved by first transforming the perspective view of the target frame into the reference view, which not only increases the accuracy of the block matching for the road surface but also improves the processing speed. The disparities are then estimated iteratively using our previously published algorithm where the search range is propagated from three estimated neighbouring disparities. Since the search range is obtained from the previous iteration, errors may occur when the propagated search range is not sufficient. Therefore, a correlation maxima verification is performed to rectify this issue, and the subpixel resolution is achieved by conducting a parabola interpolation enhancement. Furthermore, a novel disparity global refinement approach developed from the Markov Random Fields and Fast Bilateral Stereo is introduced to further improve the accuracy of the estimated disparity map, where disparities are updated iteratively by minimising the energy function that is related to their interpolated correlation polynomials. The algorithm is implemented in C language with a near real-time performance. The experimental results illustrate that the absolute error of the reconstruction varies from 0.1 mm to 3 mm.
\end{abstract}

\begin{IEEEkeywords}
3D reconstruction, road condition assessment, subpixel disparity estimation, parabola interpolation, Markov Random Fields, Fast Bilateral Stereo.
\end{IEEEkeywords}

\IEEEpeerreviewmaketitle
\section{Introduction}
\IEEEPARstart{T}HE condition assessment of asphalt and concrete civil infrastructures, e.g., bridges, tunnels and pavements, is essential to ensure their usability while still providing maximum safety for the users. It also allows the government to allocate the limited resources for maintenance and appraise long-term investment schemes \cite{Koch2015}. 
The manual visual inspections performed by either structural engineers or certified inspectors are cost-intensive, time-consuming and cumbersome \cite{Kim2014}. In 2014, a one-off investment of \pounds12bn was suggested  by the Asphalt Industry Alliance to improve the road condition across England and Wales \cite{BBC_Es_Ro_Re}.
Over the last decade, various technologies such as remote sensing, vibration sensing and computer vision have been increasingly applied in civil engineering to assess the physical and functional condition of the infrastructures such as potholes, cracking, etc.

The remote sensing methods which have been used in satellites, aeroplanes, unmanned aerial vehicles or multi-purpose survey vehicles have indeed reduced the workload of inspectors.
However, the traditional geotechnical methods can never be entirely replaced by the remote sensing approaches \cite{Schnebele2015}. Using accelerometers and GPS for data acquisition, vibration-based methods always cause distress misdetection in spite of their advantages of small storage requirements, cost-effectiveness and real-time performance \cite{Kim2014}. 
As for the approaches based on 2D computer vision, the spatial structure of the road surface cannot be illustrated explicitly \cite{Kim2014}. Therefore, 3D reconstruction-based methods are more feasible to overcome these disadvantages and simultaneously provide an enhancement in terms of detection accuracy and processing efficiency.

3D reconstruction methods can be classified as laser scanner-based, Microsoft Kinect-based and passive sensor-based. The laser scanner collects the reflected laser pulse from an object to construct its accurate 3D model \cite{Schnebele2015}. Although it provides accurate modelling results, the laser scanner equipment used for road condition analysis is still costly \cite{Kim2014}. As for the methods based on the Microsoft Kinect sensor, the depth measurement for the outdoor environment is somewhat ineffective, especially for materials which strongly absorb the infrared light \cite{Cruz2012a}. Therefore, the passive sensor-based methods, e.g., stereo vision, are more capable of reconstructing the 3D road surface for condition assessment or damage detection.

\begin{figure*}[!t]
	\begin{center}
		\centering
		\includegraphics[width=1\textwidth]{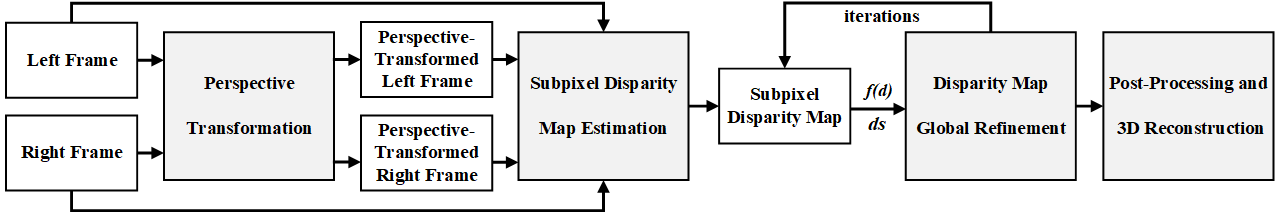}
		\caption{Stereo vision-based road surface 3D reconstruction system workflow.}
		\vspace{-1.6em}
		\label{fig.reconstruction_block_diagram}
	\end{center}
\end{figure*}

To reconstruct a real-world environment with passive sensing techniques, multiple camera views are required \cite{Hartley2003}. Images from different viewpoints can be captured using either a single moveable camera or an array of cameras \cite{Sadjadi2010}. In this paper, we use a ZED stereo camera to acquire a pair of images for road surface 3D reconstruction. Since the stereo rig is assumed to be well-calibrated, the main work performed in this paper is the disparity estimation. 
The algorithms for disparity estimation can be classified as local, global and semi-global. Local algorithms simply match a series of blocks and select the correspondence with the lowest cost or the highest correlation. This optimisation is also known as winner-take-all (WTA). 
 Unlike local algorithms, global algorithms process the stereo matching using some more sophisticated optimisation techniques, e.g., Graph Cut (GC) \cite{Boykov2001} and Belief Propagation (BP) \cite{Ihler2005}. These algorithms are commonly developed based on the Markov  Random Fields (MRF) \cite{Tappen2003}, where finding the best disparities is formulated as a probability maximisation problem. This is later addressed by energy minimisation approaches.
Semi-global matching (SGM) \cite{Hirschmuller2008} approximates the MRF inference by performing cost aggregation along all directions in the image and this greatly improves the accuracy and efficiency of stereo matching. 
 However, the occlusion problem always makes it difficult to find the optimum value for the smoothness parameters: 
 over-penalising the smoothness term can help avoid the ambiguities around discontinuities but on the other hand can lead to errors for continuous areas \cite{Mozerov2015}. 
Therefore, some authors have proposed to break down the global problem into multiple local problems, each of which is affected by uncertainties to a lesser extent \cite{Sinha2014}. For instance, one alternative way of setting smoothness parameters is to group pixels in the image into different slanted planes \cite{Bleyer2012, Yamaguchi2014, Sinha2014}. Disparities in different plane groups are estimated with local constraints. However, this results in high computational complexities, making real-time performance challenging. 

In order to further improve the trade-off between speed and accuracy, seed-and-grow local algorithms have been used extensively. In these algorithms, the disparity map is grown from a selection of seeds to minimise expensive computations and reduce mismatches caused by ambiguities.
For example, the authors of  \cite{Sara2002, Sara2006, Cech2007} presented an efficient quasi-dense stereo matching algorithm, named growing correspondence seeds (GCS), to estimate disparities iteratively with the search range propagated from a collection of reliable seeds. 
Similarly, various Delaunay triangulation-based stereo matching algorithms (DTSM) have been proposed in \cite{Spangenberg2013, Miksik2015, Pillai2016} to estimate tunable semi-dense disparity maps with the support of a piecewise planar mesh.
Our previous algorithm \cite{Zhang2013, Fan2017} also provides an efficient strategy for local stereo matching whereby the search range on row $v$ is propagated from three estimated neighbouring disparities on row $v+1$. 
Our algorithm performs better than GCS and DTSM in terms of estimating dense disparity maps for road scenes where the road disparities decrease gradually from the bottom to the top, while the disparities of obstacles remain the same.
The aim of this paper is to reconstruct the road scenes for pothole detection. In this regard, the proposed disparity estimation algorithm is developed based on our previous work in \cite{Fan2017}. To assess the condition of a road surface, millimetre accuracy is desired in 3D reconstruction and thus disparities in subpixel resolution are inevitable. Therefore, the correlation costs around the initial disparity are interpolated into a parabola and the position of the extrema is selected as the subpixel disparity.

However, the subpixel disparity maps obtained from parabola interpolation are still unsatisfactory because the correlation costs of neighbourhood systems are not aggregated before finding the best disparities. To aggregate neighbouring costs adaptively, some authors have
proposed to filter the whole cost volume with a bilateral filter since it provides a feasible solution for the initial message passing problem on a fully connected MRF \cite{Mozerov2015}. These algorithms are also known as Fast Bilateral Stereo (FBS) \cite{Yang2009, Hosni2013, Zhang2012a}. However, the intensive computational complexity introduced when filtering the whole cost volume severely impact on the processing speed.  In this regard, we believe that only the candidates around the best disparities need to be processed and a novel disparity refinement approach is proposed in this work. The workflow of our stereo vision-based road surface 3D reconstruction system is depicted in Fig. \ref{fig.reconstruction_block_diagram}.

Firstly, the perspective view of the road surface in the target image is transformed into its reference view, which greatly enhances the similarity of the road surface between the two images. Since the propagated search range is sometimes insufficient, the desirable disparities have to be further verified to ensure they possess the highest correlation costs. The latter ensures the feasibility of parabola interpolation-based subpixel enhancement. To further optimise the obtained subpixel disparity map, the interpolated parabola functions $f(d)$ are set as the labels in the MRF because they contain the information of both disparity values and correlation costs. By updating the parabola functions $f(d)$ and subpixel disparities $d_s$ iteratively, a disparity in a continuous area becomes smooth but it is preserved when discontinuities occur. 
Finally, each 3D point on the road surface is computed based on its projections on the left and right images.  The reconstruction accuracy is evaluated using  three sample models (see section \ref{sec.experimental_setup} for more details). Our datasets are publicly available at: http://www.ruirangerfan.com.

The rest of the paper is structured as follows: section \ref{sec.perspective_transformation} presents a novel perspective transformation (PT) method. 
In section \ref{sec.subpx_disparity_est}, we describe a subpixel disparity estimation algorithm. A disparity map global refinement approach is introduced in section \ref{sec.3d_global_refine}. In section \ref{sec.post_processing_3d_rec}, the disparity map is post-processed and the 3D road surface is reconstructed. In section \ref{sec.experimental_results}, the experimental results are illustrated and the performance of the proposed algorithm is evaluated. Finally, section \ref{sec.3d_conclusion} summarises the paper and provides some recommendations for future work. 

\section{Perspective Transformation}
\label{sec.perspective_transformation}

In this paper, the proposed algorithm focuses entirely on the road surface which can be treated as a ground plane (GP). To enhance the accuracy of stereo matching, we first draw on the concept of ground plane constraint in \cite{Hattori2000} and \cite{Nakai2004} to transform the perspective views of two images before estimating their disparities. GP constraint is commonly used in a wide range of obstacle detection systems, where the image on one side is set as the reference and the other image is transformed into the reference view. Pixels arising from the GP satisfy the same affine transformation while an object above the GP will not be transformed successfully \cite{Hattori2000}. Referring to the experimental results in \cite{Nakai2004}, pixels from an obstacle are distorted in the transformed image. Nevertheless, the GP in the transformed image looks more similar to its reference view. Therefore, a perspective transformation makes the obstacle areas noisy and unreliable but greatly enhances the similarity of the road surface between two images.  In this paper, the road surface is defined as: 
\begin{equation}
\boldsymbol{n}^\top \boldsymbol{P_w}+\beta=0
\label{eq.road_surface_func}
\end{equation}
where $\boldsymbol{P_w}=[X_w,Y_w,Z_w]^\top$ is an arbitrary 3D point on the road surface. Its projections on the left image $\pi_l$ and the right image $\pi_r$ are $\boldsymbol{p_l}=[u_l,v_l]^\top$ and $\boldsymbol{p_r}=[u_r,v_r]^\top$, respectively. $\boldsymbol{n}=[n_0, n_1, n_2]^\top$ is the normal vector of the road surface. The planar transformation between $\boldsymbol{\tilde{p}_l}=[u_l,v_l,1]^\top$ and $\boldsymbol{\tilde{p}_r}=[u_r,v_r,1]^\top$ is given in Eq. \ref{eq.homograph_mat} \cite{Hartley2003}. Here, $\boldsymbol{\tilde{p}}=[u,v,1]^\top$ denotes the homogeneous coordinate of $\boldsymbol{p}=[u,v]^\top$.

\begin{equation}
\boldsymbol{\tilde{p}_r}=\boldsymbol{H_{rl}}\boldsymbol{\tilde{p}_l}
\label{eq.homograph_mat}
\end{equation}

$\boldsymbol{H_{rl}}\in\mathbb{R}^{3\times3}$ denotes a homograph matrix, which is generally used to distinguish obstacles from the road surface \cite{Hattori2000}. It can be decomposed as \cite{Hartley2003}: 

\begin{equation}
\boldsymbol{H_{rl}}=
\boldsymbol{K_r} 
\left( \boldsymbol{R_{rl}}-\frac{\boldsymbol{T_{rl}}\boldsymbol{n}^\top}{\beta}
\right)
\boldsymbol{K_l}^{-1}
\label{eq.H_dec}
\end{equation}
where $\boldsymbol{R_{rl}}$ is a SO(3) matrix and $\boldsymbol{T_{rl}}$ is a translation vector.  $\boldsymbol{P_l}$ in the left camera coordinate system can be transformed to $\boldsymbol{P_r}$ in the right camera coordinate system according to $\boldsymbol{P_r}=\boldsymbol{R_{rl}}\boldsymbol{P_l}+\boldsymbol{T_{rl}}$.  $\boldsymbol{K_l}$ and $\boldsymbol{K_r}$ are intrinsic matrices of the two cameras. For a well-calibrated stereo system, $\boldsymbol{R_{rl}}$, $\boldsymbol{T_{rl}}$, $\boldsymbol{K_l}$ and $\boldsymbol{K_r}$ are already known. We only need to estimate $\boldsymbol{n}$ and $\beta$ for $\boldsymbol{H_{rl}}$. Generally, $\boldsymbol{H_{rl}}$ can be estimated with at least four pairs of correspondences $\boldsymbol{{p}_l}$ and $\boldsymbol{{p}_r}$ \cite{Hartley2003}. Hattori et al. proposed a pseudo-projective camera model where several assumptions are made about road geometry to simplify the estimation of $\boldsymbol{H_{rl}}$ \cite{Hattori2000}. In this paper, we improve on their algorithm by considering the following hypotheses:

\begin{itemize} 
	\item $\boldsymbol{K_l}$ and $\boldsymbol{K_r}$ are identical.
	\item $\boldsymbol{R_{rl}}$ is an identity matrix.
	\item $\boldsymbol{T_{rl}}$ is in the same direction as the $X_w$-axis.
	\item the road surface is a horizontal plane: $n_1Y_w+\beta=0$.
	\item rotation of the stereo rig is only about the $X_w$-axis. 
\end{itemize}

For a perfectly-calibrated stereo rig, $v_l=v_r=v$. The disparity is defined as $d=u_l-u_r$. The projection of a horizontal plane on the v-disparity map is a linear pattern \cite{Hu2005}:
\begin{equation}
d=-\frac{T_cn_1}{\beta}(f\sin\theta-v_0\cos\theta)-v\frac{T_cn_1}{\beta}\cos\theta=\alpha_0+\alpha_1 v
\label{eq.relation_vd}
\end{equation}
where $\theta$ is the pitch angle between the stereo rig and the road surface (an example can be seen in Fig. \ref{fig.extrinsic_rotations} (a)), $f$ is the focus length of the cameras, $T_c$ is the baseline, and $(u_0,v_0)$ is the principal point in pixels. When $\theta=\pi/2$, $d=-{fT_cn_1}/{\beta}$ is a constant. Otherwise, $d$ is proportional to $v$ \cite{Hu2005}. This implies that a perspective distortion always exists for the GP in two images, which further affects the accuracy of block matching. Therefore, the PT aims to make the GP in the transformed image similar to that in the reference frame. 

 \begin{figure}[!t]
	\centering
	\includegraphics[width=0.48\textwidth]{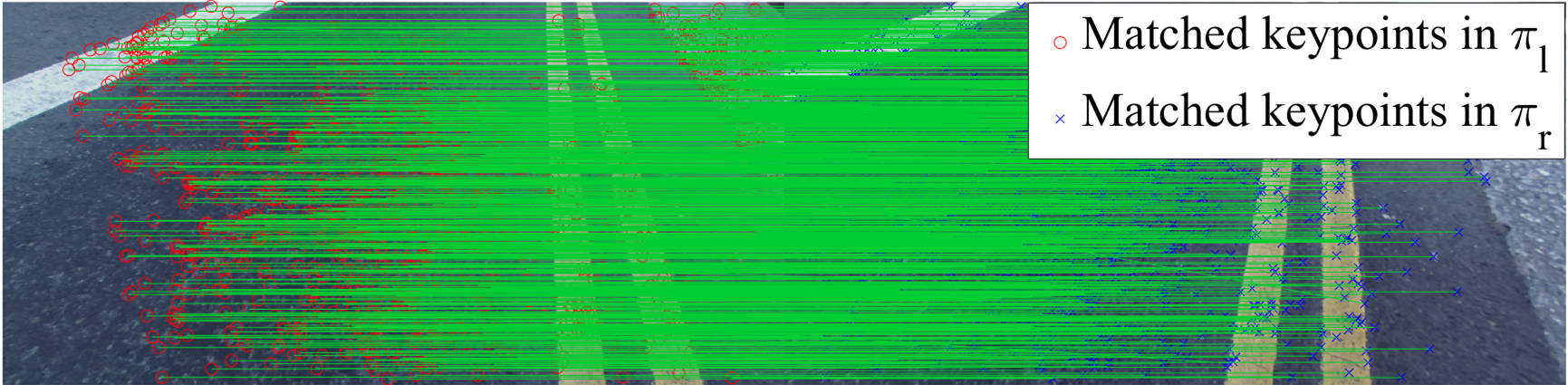}
	\caption{BRISK-based on-road keypoints detection and matching between the left and right images.}
	\vspace{-1.35em}
	\label{fig.brisk}
\end{figure}

Now, the PT can be straightforwardly realised using parameters $\boldsymbol{\alpha}=[\alpha_0,\alpha_1]^\top$. The proposed PT is detailed in algorithm \ref{al.pt}. $\boldsymbol{\alpha}$ can be estimated by solving a least squares problem with a set of reliable correspondences $\boldsymbol{Q_l}=[\boldsymbol{{p}_{l1}},\boldsymbol{{p}_{l2}},\dots,\boldsymbol{{p}_{lm}}]^\top$ and $\boldsymbol{Q_r}=[\boldsymbol{{p}_{r1}},\boldsymbol{{p}_{r2}},\dots,\boldsymbol{{p}_{rm}}]^\top$. In this paper, we use BRISK (Binary Robust Invariant Scalable Keypoints) to detect and match $\boldsymbol{Q_l}$ and $\boldsymbol{Q_r}$. It allows a faster execution to achieve approximately the same number of correspondences as SIFT (Scale-Invariant Feature Transform) and SURF (Speeded-Up Robust Features) \cite{Leu2011}. An example of on-road keypoints detection and matching is illustrated in Fig. \ref{fig.brisk}.

 \begin{algorithm}[h!]
	\KwData{$\pi_l$ and $\pi_r$}
	\KwResult{$\boldsymbol{\alpha}=[\alpha_0,\alpha_1]^\top$}
	detect and match the keypoints in $\pi_l$ and $\pi_r$\;
    \uIf{$|v_{li}-v_{ri}|>\epsilon$ or $u_{li}-u_{ri}<0$}
	{
		remove $\boldsymbol{{p}_{li}}$ and $\boldsymbol{{p}_{ri}}$ from $\boldsymbol{Q_l}$ and $\boldsymbol{Q_r}$, respectively\; 
	}
	estimate $\boldsymbol{\alpha}$ using the least squares fitting\;
	all points in the target image are shifted $\alpha_0+\alpha_1v-\delta$ pixels to the reference view\;
	\caption{Perspective transformation.}
	\label{al.pt}
\end{algorithm}

\begin{figure}[!t]
	\centering
	\subfigure[]
	{
		\includegraphics[width=0.225\textwidth]{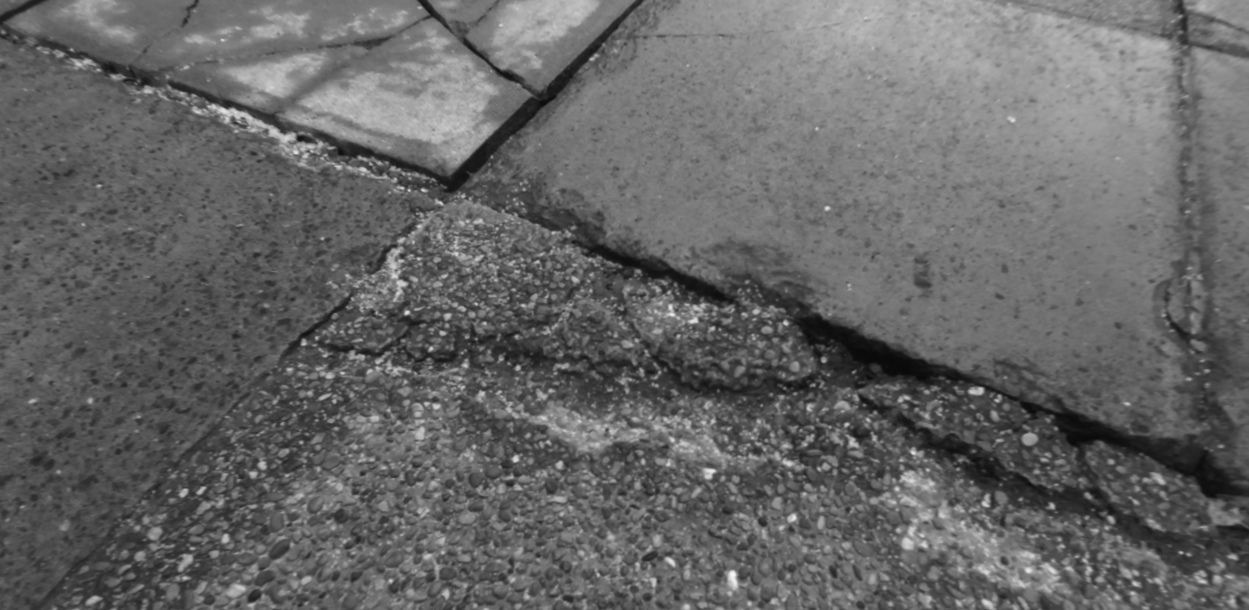}
	}
	\subfigure[]
	{
		\includegraphics[width=0.225\textwidth]{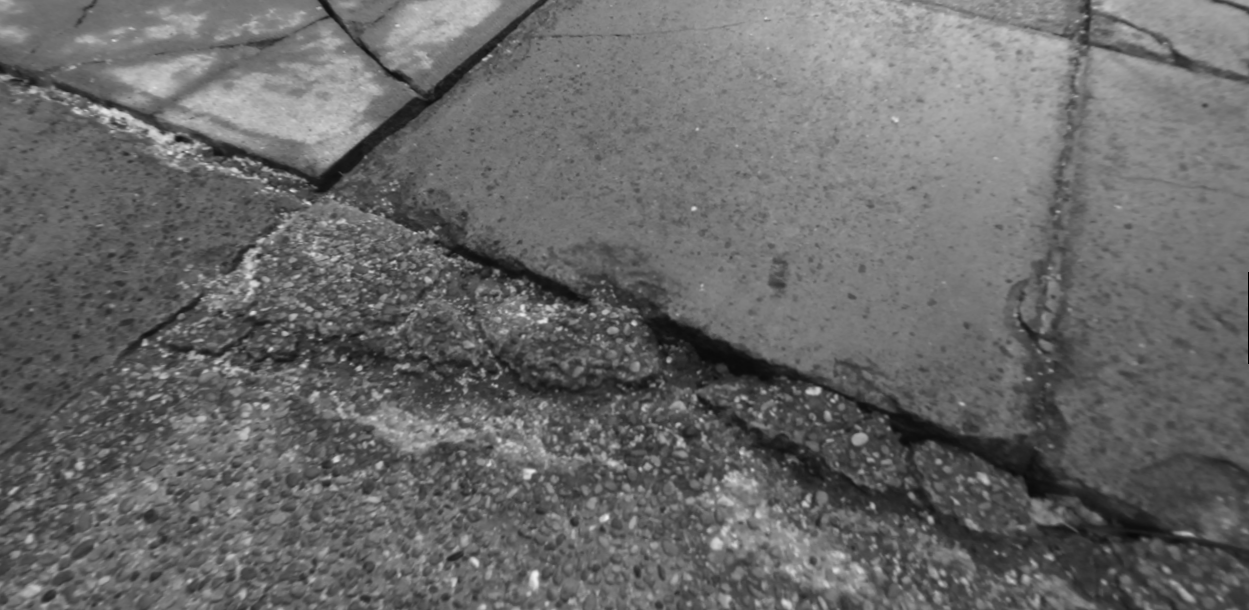}
	}
	\subfigure[]
	{
		\includegraphics[width=0.225\textwidth]{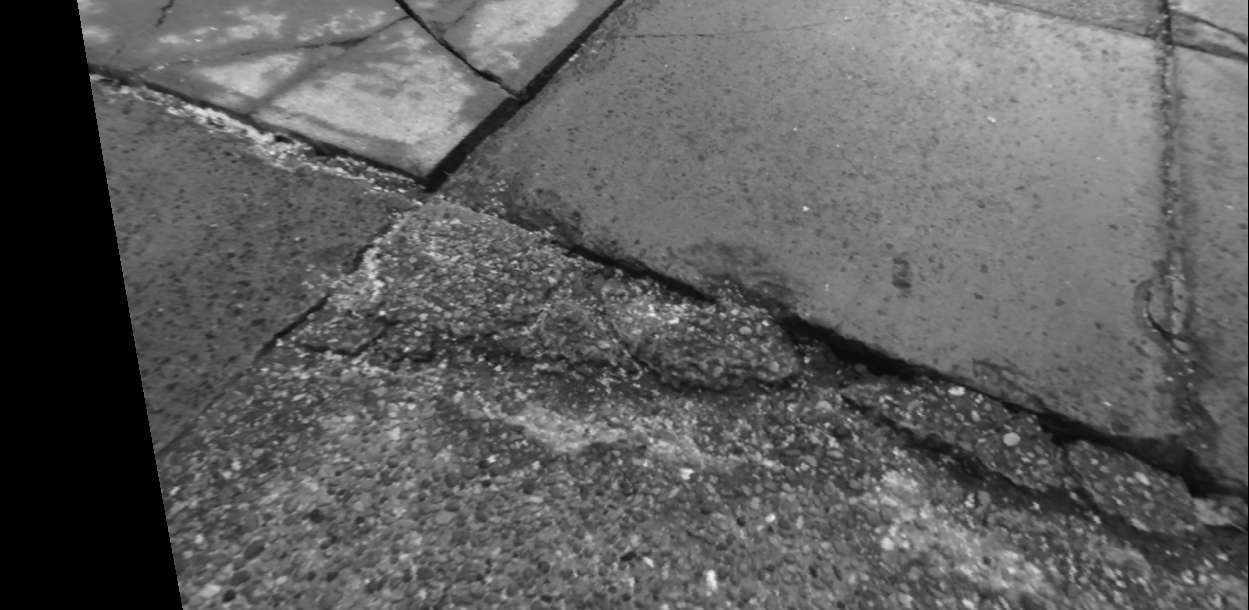}
	}
	\subfigure[]
	{
		\includegraphics[width=0.225\textwidth]{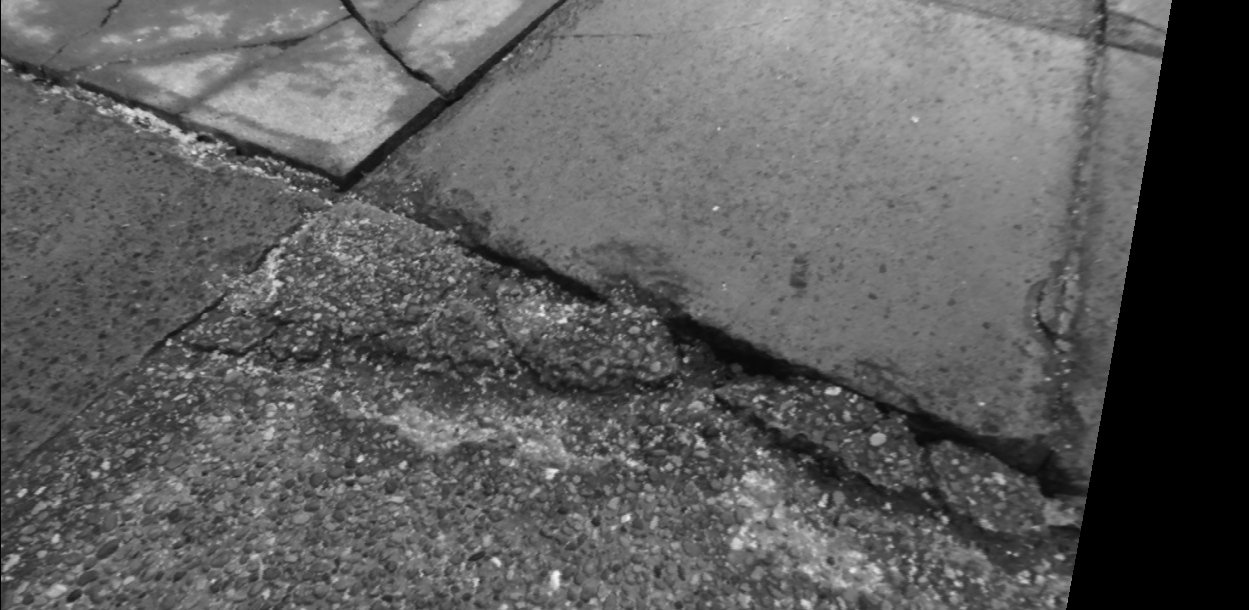}
	}
	\caption{Perspective transformation. (a) left image. (b) right image. (c) transformed right image. (d) transformed left image. (a) and (c) are used as the input left and right images for the left disparity map estimation. (d) and (b) are used as the input left and right images for the right disparity map estimation.}
		\vspace{-1.15em}
	\label{fig.pt_results}
\end{figure}

Since outliers can severely affect the accuracy of least squares fitting, we first remove the less reliable correspondences before estimating $\boldsymbol{\alpha}$, where $\epsilon$ is proposed to be 1. For the left disparity map $\ell^{lf}$ estimation, each point on row $v$ in $\pi_r$ is shifted $\alpha_0+\alpha_1v-\delta$ pixels to the right, where $\delta$ is a constant set to 20 (for dataset 1 and 2) or 30 (for dataset 3) to guarantee that all the disparities are positive. Similarly, each point in $\pi_l$ is shifted $\alpha_0+\alpha_1v-\delta$ pixels to the left when $\pi_r$ is served as the reference. An example of perspective transformation is presented in Fig. \ref{fig.pt_results}. The performance improvements achieved by using the PT will be discussed in section \ref{sec.experimental_results}.

\section{Subpixel disparity map estimation}
\label{sec.subpx_disparity_est}

As compared to many other stereo matching algorithms which aim at automotive applications, the trade-off between speed and precision has been greatly improved in our previous work \cite{Zhang2013} and \cite{ Fan2017}. The subpixel accuracy can be achieved by conducting a parabola interpolation for the correlation costs around the initial disparity \cite{Yang2009}. The subpixel disparity global refinement will be discussed in section \ref{sec.3d_global_refine}.

\subsection{Stereo Matching}

In this paper, our previous algorithm \cite{Zhang2013} is utilised to estimate integer disparities, where the NCC (Normalised Cross-Correlation) is used to compute the matching costs, and the search range $SR$ for pixel at $(u,v)$ is propagated from three estimated neighbouring disparities on row $v+1$. To accelerate the NCC execution, we rearrange the NCC equation as follows:

\begin{equation}
c(u,v,d)=
\frac{1}{n\sigma_{l}\sigma_{r}}
\Bigg(
{\sum\limits_{x=u-\rho}^{x=u+\rho}\sum\limits_{y=v-\rho}^{y=v+\rho}i_{l}(x,y)i_{r}(x-d,y)
-{n\mu_{l}}\mu_{r}}
\Bigg)
\label{eq.ncc_developed}
\end{equation}
where $c(u,v,d)$ is defined as the correlation cost between two square blocks selected from $\pi_l$ and $\pi_r$, and a higher $c(u,v,d)$ corresponds to a better matching and vice-versa. $i_l$ or $i_r$ is the intensity of a pixel in $\pi_l$ or $\pi_r$. The edge length of the square block is $2\rho+1$, and $n$ represents the number of pixels in it. $(u,v)$ and $(u-d,v)$ are the centres of the left and right blocks, respectively. $\mu{_{l}}$ and $\mu{_{r}}$ denote the means of the intensities within the two blocks. $\sigma_{l}$ and $\sigma_{r}$ are their standard deviations. 

From Eq. \ref{eq.ncc_developed}, $\mu$ and $\sigma$ only matter for each independent block selected from $\pi_l$ or $\pi_r$, and $d$ determines a pair of blocks for matching. Therefore, the calculation of $\mu_l$, $\mu_r$, $\sigma_{l}$ and $\sigma_{r}$ will always be repeated in conventional NCC-based stereo matching algorithms. In \cite{Fan2017}, we propose to pre-calculate the values of $\mu$ and $\sigma$ and store them in a static program storage for direct indexing. Thus, the computational complexity of the NCC is simplified to a dot product, making stereo matching more efficient. More details on the implementation procedure are available in \cite{Fan2017}.

\subsubsection{Search Range Propagation (SRP)}
\begin{figure}[!b]
	\centering
	\centering
	\subfigure[]
	{
		\includegraphics[width=0.225\textwidth]{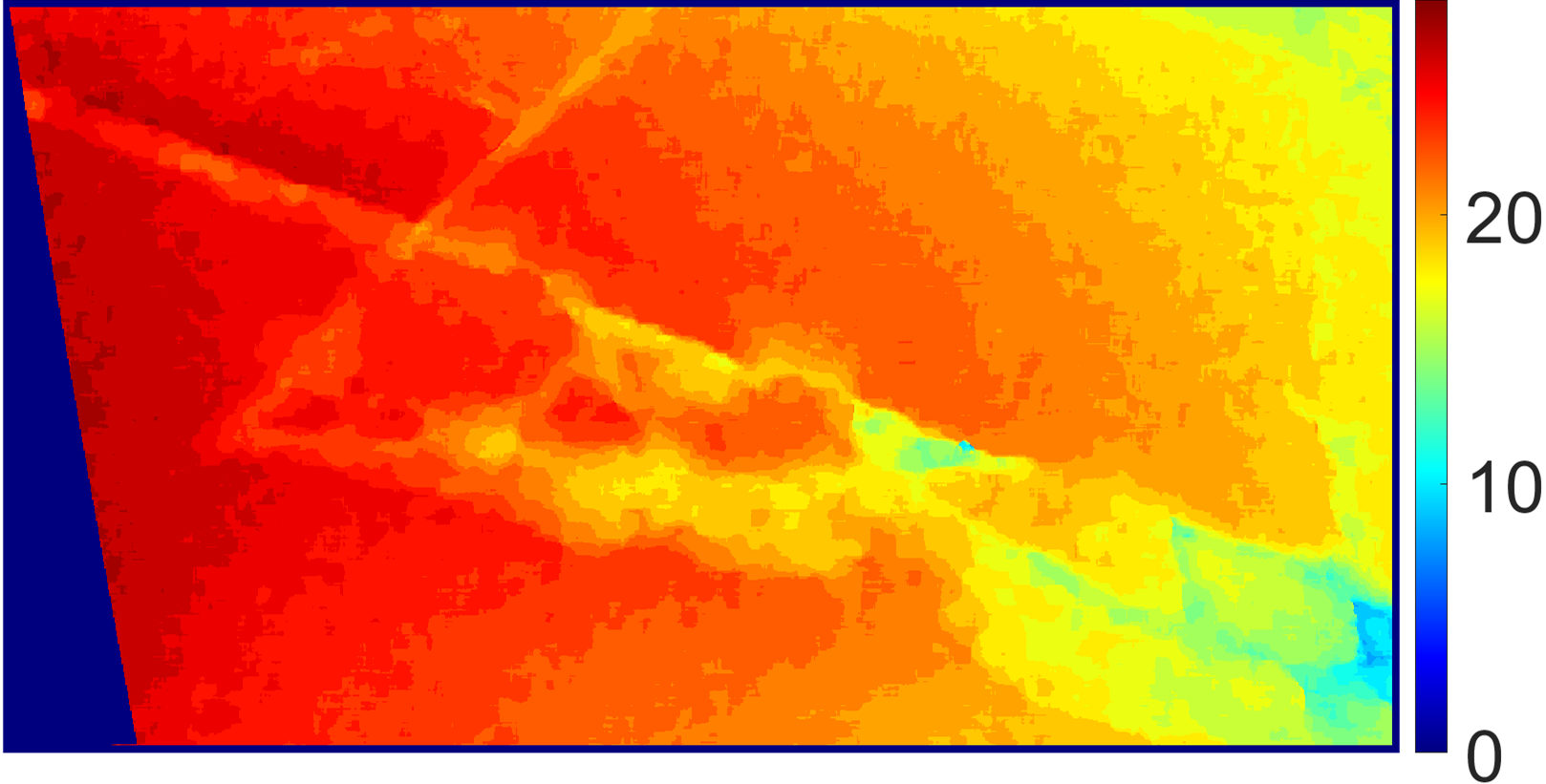}
	}
	\subfigure[]
	{
		\includegraphics[width=0.223\textwidth]{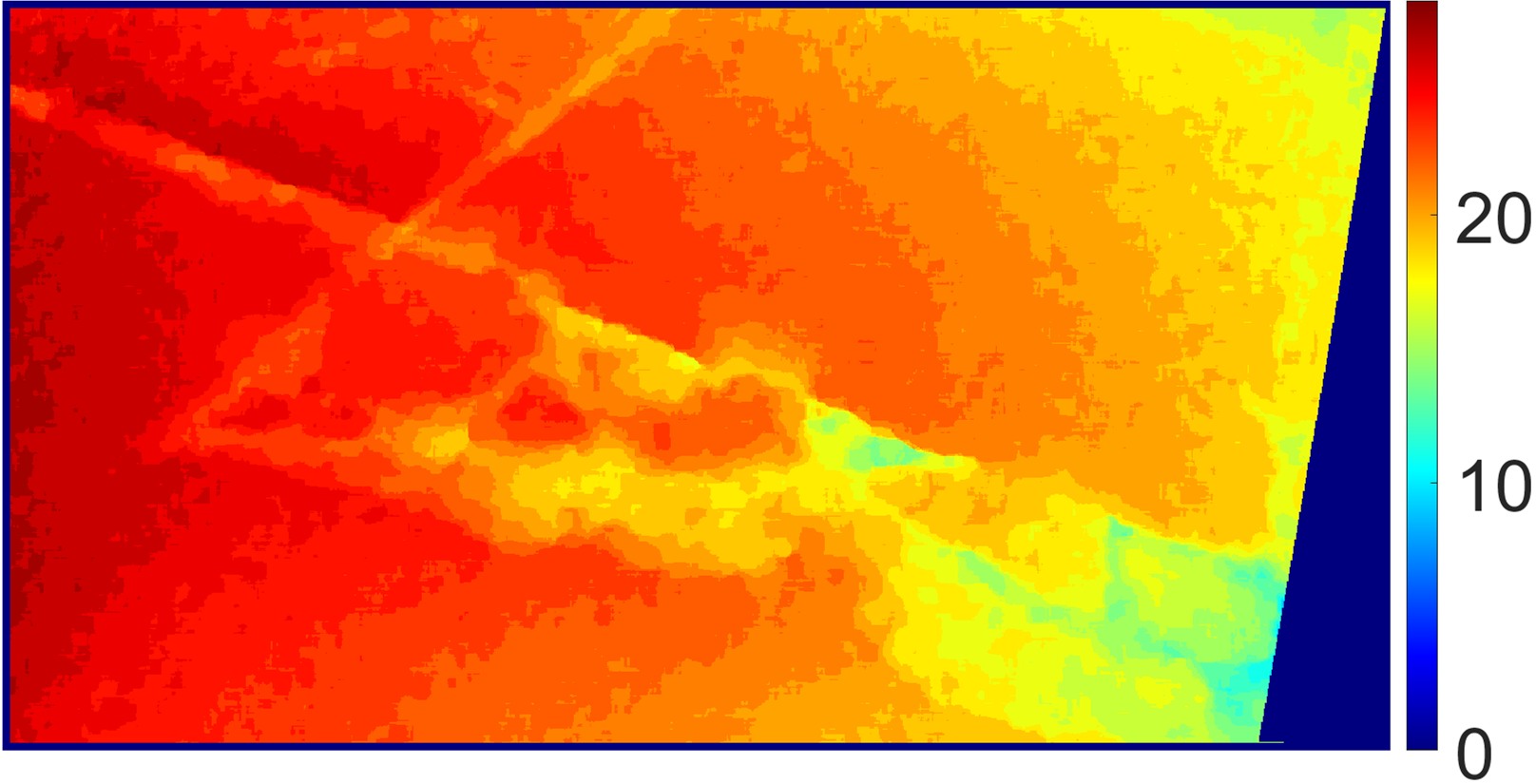}
	}
	\subfigure[]
	{
		\includegraphics[width=0.225\textwidth]{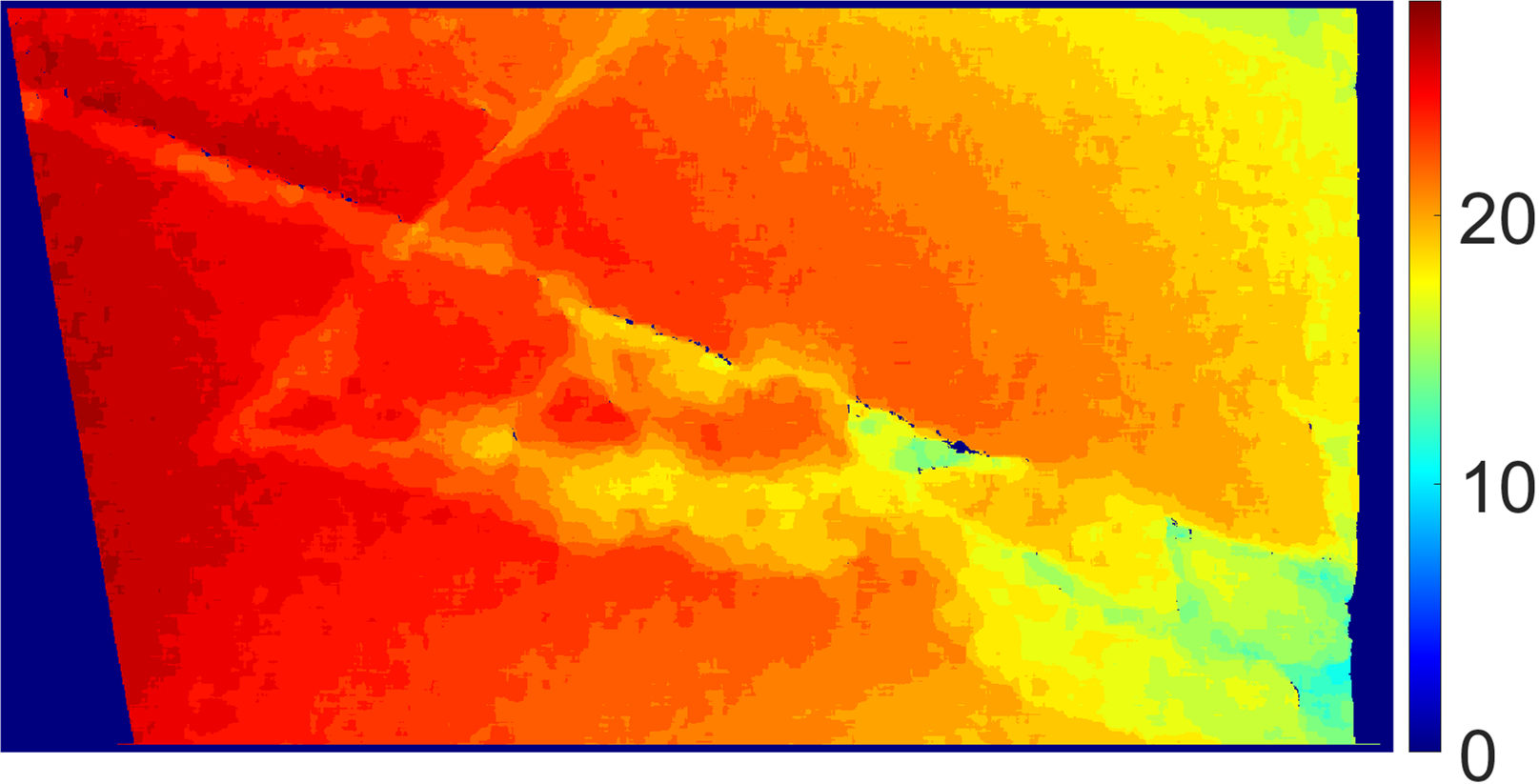}
	}
	\subfigure[]
	{
		\includegraphics[width=0.225\textwidth]{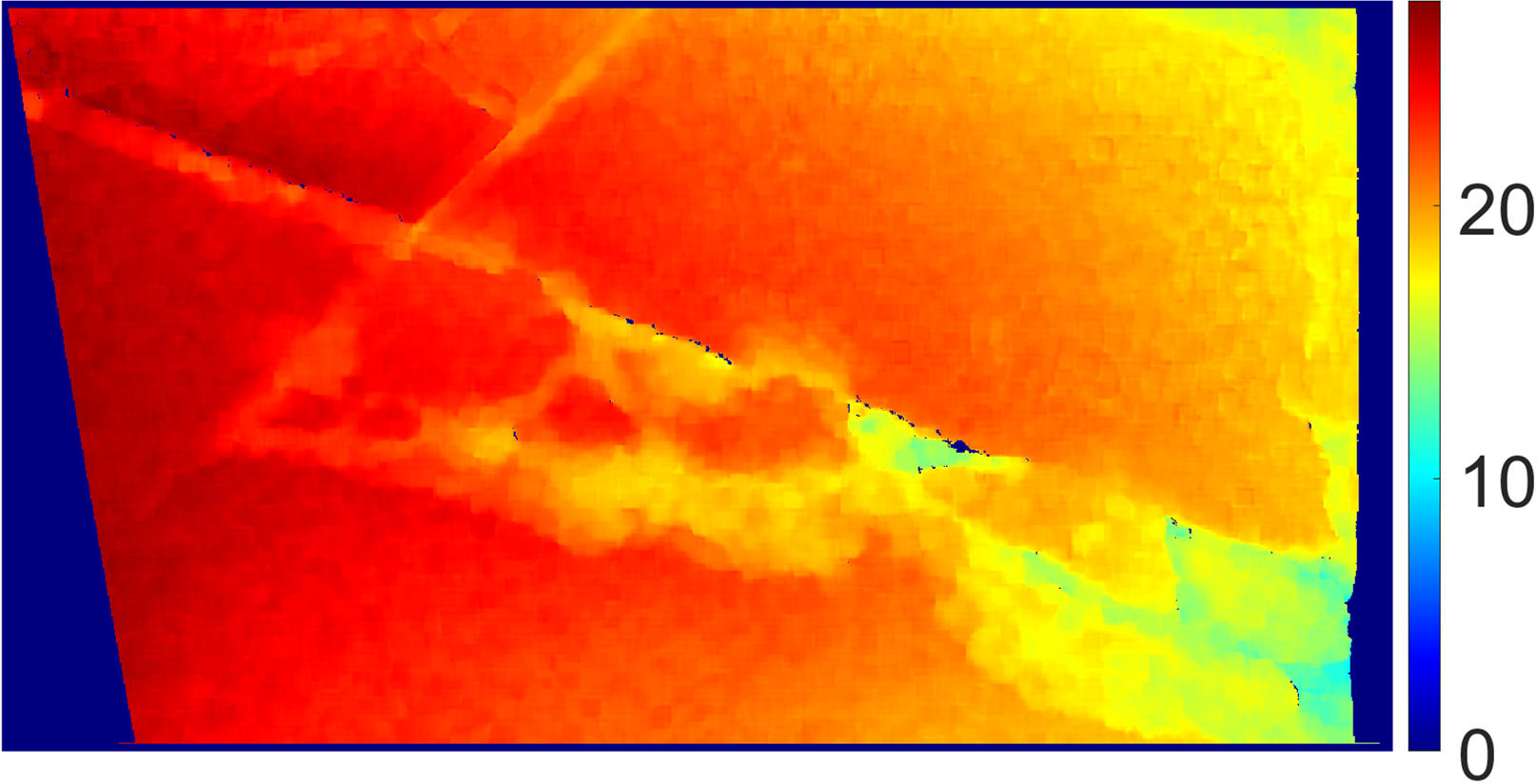}
	}
	\caption{Subpixel disparity map estimation. (a) left disparity map. (b) right disparity map. (c) left disparity map processed with the LRC check. (d) subpixel disparity map. }
	\label{fig.subpixel_disparity_map}
\end{figure}

Since the concept of "local coherence constraint" was proposed in \cite{Roy1999}, many researchers have turned their focus on seed-and-grow algorithms for stereo matching. Either semi-dense or quasi-dense disparity maps can be estimated efficiently with the guidance from a collection of reliable feature points \cite{Sara2002, Sara2006, Cech2007, Spangenberg2013, Miksik2015, Pillai2016}. In this paper, the road surface is treated as a GP whose disparities change gradually from the bottom of the image to its top, which makes our previous algorithm \cite{Zhang2013} more efficient than other methods in terms of estimating an accurate dense disparity map.  
The proposed algorithm propagates the search range $SR$ iteratively row by row from the bottom of the image to its top. In the first iteration, the disparity estimation performs a full search range. 
 Then, $SR$ at $(u,v)$ is propagated from three estimated neighbouring disparities 
using Eq. \ref{eq.srp}, where $\tau$ is the bound of $SR$ and is set as 1 in this paper. The left and right disparity maps, $\ell^{lf}$ and $\ell^{rt}$, are shown in Fig. \ref{fig.subpixel_disparity_map} (a) and (b), respectively.

\begin{equation}
SR={\bigcup_{k=u-1}^{u+1}}\{sr|sr\in[\ell(k,v+1)-\tau,\ell(k,v+1)+\tau]\}
\label{eq.srp}
\end{equation}
\subsubsection{Correlation Maxima Verification (CMV)}
Since the search range propagates using Eq. \ref{eq.srp}, errors may occur in subpixel enhancement when $c(u,v,d-1)$ or $c(u,v,d+1)$ is not computed and compared with $c(u,v,d)$. 
Therefore, CMV will run until the correlation cost of the disparity is a local maxima. More details are provided in algorithm \ref{al.cmv}.

  \begin{algorithm}[!h]
	\KwData{disparity map $\ell$}
	\KwResult{correlation maxima verified disparity map $\ell_{cmv}$}
	\uIf{$c(u,v,d)>\max\{c(u,v,d-1),c(u,v,d+1)\}$}{
		$\ell_{cmv}(u,v)\gets \ell(u,v)$\;
	}
	\uElseIf{$c(u,v,d-1)<c(u,v,d)<c(u,v,d+1)$}{
		\Repeat{$c(u,v,d+k)<c(u,v,d+k-1)$}{
			compute $c(u,v,d+k),\ k\ge2$;
		}
		$\ell_{cmv}(u,v)\gets d+k-1$\;
	}
	\Else{
		\Repeat{$c(u,v,d-k)<c(u,v,d-k+1)$}{
			compute $c(u,v,d-k),\  k\ge2$;
		}
		$\ell_{cmv}(u,v)\gets d-k+1$\;
	}
	\caption{Correlation maxima verification}
	\label{al.cmv}%
\end{algorithm}

\vspace{-1.6em}

\subsection{Left-Right Consistency (LRC) Check}
Due to the fact that each pair of correspondences from two images is unique, if we select an arbitrary pixel $(u,v)$ from the left disparity map $\ell^{lf}$, there should exist at most one correspondence in the right disparity map $\ell^{rt}$ \cite{Mozerov2015}:

\begin{equation}
\ell^{lf}(u,v)=\ell^{rt}(u-\ell^{lf}(u,v),v)
\label{eq.lrc}
\end{equation}

 Pixels that are only visible in one disparity map are marked as uncertainties. A LRC check is performed to remove these half-occluded areas. Although the LRC check doubles the computational complexity by re-projecting the estimated disparities from one disparity map to the other one, most of the infeasible conjugate pairs can be removed, and an outlier in the disparity map can be found. The left disparity map after the LRC check processing is illustrated in Fig. \ref{fig.subpixel_disparity_map} (c).

\subsection{Subpixel Enhancement}
\label{sec.subpixel_enhancement}
In this paper, the road surface application requires a millimetre accuracy in 3D reconstruction. A disparity error larger than one pixel may result in a non-neglected difference in the reconstructed road surface \cite{Haller2012}. Therefore, subpixel resolution is inevitable to achieve a highly accurate result. 

For each pixel whose disparity $d$ is $\ell(u,v)$, we fit a parabola to three correlation costs $ c(u,v,d-1)$, $c(u,v,d)$ and $c(u,v,d+1)$ around the initial disparity $d$. The centreline of the parabola is selected as the subpixel displacement $d_s$ as follows \cite{Zhang2012a}:
\begin{equation}
d_s=d+\frac{c(u,v,d-1)-c(u,v,d+1)}{2c(u,v,d-1)+2c(u,v,d+1)-4c(u,v,d)}
\label{eq.subpixel_disparity}
\end{equation}

Since the CMV guarantees that $c(u,v,d)$ is larger than both $c(u,v,d-1)$ and $c(u,v,d+1)$, $d_s$ will be between $d-1$ and $d+1$. Fig. \ref{fig.subpixel_disparity_map} (c) after the subpixel enhancement is given in Fig. \ref{fig.subpixel_disparity_map} (d). 

\section{Disparity Map Global Refinement}
\label{sec.3d_global_refine}

\subsection{Markov Random Fields and Fast Bilateral Stereo}
\label{sec.mrf_fbs}
Unlike the principle of WTA applied in local stereo matching algorithms, the matching costs from neighbouring pixels are also taken into account in global algorithms, e.g., GC and BP. The MRF is a commonly used graphical model in these global algorithms. An example of the MRF model is depicted in Fig. \ref{fig.mrf}.

The graph  $\mathcal{G}=(\mathcal{P},\mathcal{E})$ is a set of vertices $\mathcal{P}$ connected by edges $\mathcal{E}$, where $\mathcal{P}=\{\boldsymbol{p_{11}},\boldsymbol{p_{12}},\cdots,\boldsymbol{p_{mn}}\}$ and $\mathcal{E}=\{(\boldsymbol{p_{ij}},\boldsymbol{p_{st}})\ |\ \boldsymbol{p_{ij}},\boldsymbol{p_{st}}\in\mathcal{P}\}$.  
Two edges sharing one common vertex are called a pair of adjacent edges \cite{Blake2011}. Since the MRF is considered to be undirected,  $(\boldsymbol{p_{ij}},\boldsymbol{p_{st}})$ and $(\boldsymbol{p_{st}},\boldsymbol{p_{ij}})$ refer to the same edge here. $\mathcal{N}_{ij}=\{\boldsymbol{n_{1{p_{ij}}}},\boldsymbol{n_{2{p_{ij}}}},\cdots,\boldsymbol{n_{k{p_{ij}}}}\ |\ \boldsymbol{n_{{p_{ij}}}}\in\mathcal{P}\}$ is a neighbourhood system for $\boldsymbol{p_{ij}}$.

For stereo vision problems, $\mathcal{P}$ is a $m\times n$ disparity map and $\boldsymbol{p_{ij}}$ is a vertex (or node) at the site of $(i,j)$ with a label of disparity $d_{ij}$.  Because more candidates taken into consideration usually make the inference of a true disparity intractable, only the neighbours adjacent to $\boldsymbol{p_{ij}}$ are considered for stereo matching \cite{Tappen2003}. This is also known as a pairwise MRF. In this paper, $k=4$ and $\mathcal{N}$ is a four-connected neighbourhood system. $\mathcal{E}_1=(\boldsymbol{p_{ij}},\boldsymbol{n_{1{p_{ij}}}})$, $\mathcal{E}_2=(\boldsymbol{p_{ij}},\boldsymbol{n_{2{p_{ij}}}})$, $\mathcal{E}_3=(\boldsymbol{p_{ij}},\boldsymbol{n_{3{p_{ij}}}})$ and $\mathcal{E}_4=(\boldsymbol{p_{ij}},\boldsymbol{n_{4{p_{ij}}}})$ are adjacent edges sharing the vertex $\boldsymbol{p_{ij}}$. 
The disparity of $\boldsymbol{p_{ij}}$ tends to have a strong correlation with its vicinities, while it is linked implicitly to any other random nodes in the disparity map.
In  \cite{Tappen2003}, the joint probability of the MRF is written as:

\begin{equation}
P(\boldsymbol{p}, q)=\prod_{\boldsymbol{p_{ij}}\in\mathcal{P}} \Phi(\boldsymbol{p_{ij}}, q_{\boldsymbol{p_{ij}}})  
\prod_{\boldsymbol{n_{p_{ij}}}\in\mathcal{N}_{ij}} \Psi (\boldsymbol{p_{ij}}, \boldsymbol{n_{p_{ij}}})
\label{eq.mrf_eq2}   
\end{equation}
where $q_{\boldsymbol{p_{ij}}}$ represents the intensity differences, $\Phi(\cdot)$ expresses the compatibility between  possible disparities and the corresponding intensity differences, and $\Psi(\cdot)$ expresses the compatibility between $\boldsymbol{p_{ij}}$ and its neighbourhood system. Now, the aim of finding the best disparity is equivalent to maximising the probability in Eq. \ref{eq.mrf_eq2}. This can be realised by formulating Eq. \ref{eq.mrf_eq2} as an energy function \cite{Tappen2003}:

\begin{figure}[!t]
	\centering
	\includegraphics[width=0.30\textwidth]{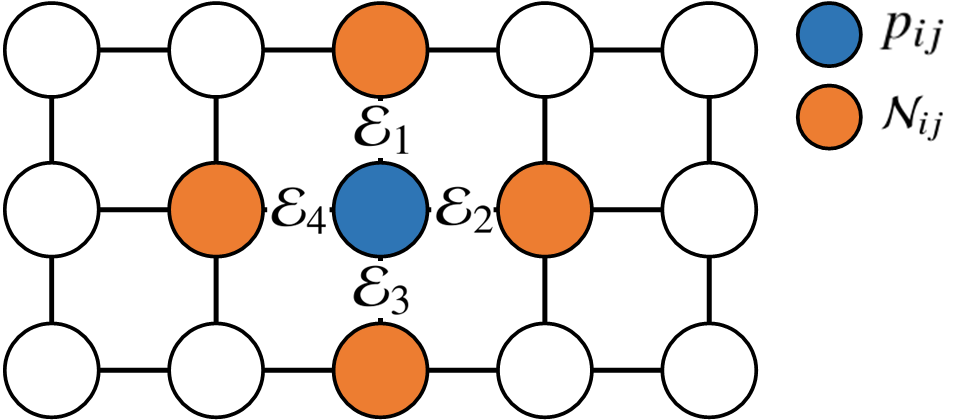}
	\caption{Markov random fields.}
	\label{fig.mrf}
	\vspace{-1.5em}
\end{figure}

\begin{equation}
\begin{split}
E(\boldsymbol{p})&=\sum_{\boldsymbol{p_{ij}}\in\mathcal{P}} D(\boldsymbol{p_{ij}}, q_{\boldsymbol{p_{ij}}})+
\sum_{\boldsymbol{n_{p_{ij}}}\in\mathcal{N}_{ij}} V (\boldsymbol{p_{ij}}, \boldsymbol{n_{p_{ij}}})\\
\end{split}
\label{eq.mrf_eq3}   
\end{equation}

$D(\cdot)$ and $V(\cdot)$ are two energy functions. $D(\cdot)$ corresponds to the matching cost and $V(\cdot)$ determines           the aggregation from the neighbours. In the MRF model, the method to formulate an adaptive $V(\cdot)$ is important because the intensity in discontinuous areas usually varies greatly from that of its neighbours \cite{Li2012}. 
 Since Tomasi et al. introduced the bilateral filter in \cite{Tomasi1998}, many authors have investigated its applications to aggregate the matching costs \cite{Yang2009, Hosni2013, Zhang2012a}. These methods are also grouped into fast bilateral stereo, where both  intensity difference and spatial distance provide a weight to adaptively constrain the aggregation of discontinuities. A general representation of the cost aggregation in FBS is represented as follows:

\begin{equation}
c_{agg}(i,j,d)=\frac{\sum_{x=i-\rho}^{i+\rho}  \sum_{y=j-\rho}^{j+\rho}  \omega_d(x,y)\omega_r(x,y)c(x,y,d)}{\sum_{x=i-\rho}^{i+\rho}  \sum_{y=j-\rho}^{j+\rho} \omega_d(x,y)\omega_r(x,y)}
\label{eq.fbs}
\end{equation}
where $\omega_d$ is based on the spatial distance and $\omega_r$ is based upon the colour similarity. The costs $c$ within a square block are aggregated adaptively to obtain $c_{agg}$.

Although the FBS has shown a good performance in terms of matching accuracy, it usually takes a long time to process the whole cost volume. Therefore, we propose an improved adaptive aggregation method to optimise the subpixel disparity map iteratively. 

\subsection{Subpixel Disparity Refinement with Energy Minimisation}
\label{sec.subpixel_disparity_refinemetn}

In this paper, the local algorithm proposed in section \ref{sec.subpx_disparity_est} greatly minimises the trade-off between accuracy and speed. A precise subpixel disparity map can be estimated with a near real-time performance. Compared to conventional MRF-based algorithms, our global refinement method only aggregates the costs around the best disparity and updates the disparity map in a more efficient way. The proposed disparity refinement algorithm is developed based on the following assumptions: 

\begin{itemize}
	\item the subpixel disparity map obtained in section \ref{sec.subpx_disparity_est} is acceptable.
	\item for an arbitrary pixel, its neighbours (excluding discontinuities) in all directions have similar disparities.
    \item the interpolated parabola $f(d)=\beta_0+\beta_1 d+\beta_2 d^2$ in section \ref{sec.subpixel_enhancement} is locally smooth.
\end{itemize}

Before going into further details about our disparity refinement approach, we first rewrite the energy function in Eq. \ref{eq.mrf_eq3} in a more general way as follows \cite{Szeliski2008}:  

\begin{equation}
E(\boldsymbol{p})=E_{data}(\boldsymbol{p_{ij}})+\lambda E_{smooth}(\boldsymbol{p_{ij}},  \boldsymbol{n_{p_{ij}}})
\label{eq.energy_minimisation}
\end{equation}
where the term $E_{data}$ penalises the solutions that are inconsistent with the observed data, $E_{smooth}$ enforces the piecewise smoothness and $\lambda$ is the smoothness parameter. For conventional MRF-based stereo matching algorithms, $E_{data}$ denotes the matching cost and $E_{smooth}$ is the cost aggregation from the neighbourhood system. By minimising the global energy of the whole random field, a disparity map can be estimated. 

In section \ref{sec.subpixel_enhancement},  we fit a parabola $f(d)=\beta_0+\beta_1 d+\beta_2 d^2$ to three correlation costs $c(u,v,d-1)$, $c(u,v,d)$ and $c(u,v,d+1)$ to get the subpixel disparity $d_s$. The parabola function $f(d)$ contains the information of both subpixel disparity and correlation costs. Since $f(d)$ is assumed to be locally smooth, the neighbouring pixels tend to have similar parabola parameters. However, when an abrupt change occurs, they vary significantly and in this case, the condition for uniform smoothness is no longer valid. 
Therefore, we use function $f(d_{\boldsymbol{p_{ij}}})$ as the label in MRF. By adaptively  aggregating functions $f(d_{\boldsymbol{n_{p_{ij}}}})$ of the neighbourhood system to $f(d_{\boldsymbol{{p_{ij}}}})$, $f(d_{\boldsymbol{{p_{ij}}}})$ is updated iteratively.

In order to ensure energy minimisation rather than energy maximisation as widely presented in literature, the term $E_{data}$ is defined as:

\begin{equation}
E_{data}(\boldsymbol{p_{ij}})=-f{({d}_{\boldsymbol{p_{ij}}})}
\label{eq.edata}
\end{equation}

$\lambda$ has a value of $1/\sqrt{2}$ in this paper. Using the same strategy of adaptive aggregation in FBS, we define the smoothness energy $E_{smooth}(\boldsymbol{p_{ij}},  \boldsymbol{n_{p_{ij}}})$ as the adaptive sum of negative interpolated parabolas $-f({d}_{\boldsymbol{n_{{p_{ij}}}}})$ of spatially varying horizontal and vertical nearest neighbours:

\begin{equation}
E_{smooth}(\boldsymbol{p_{ij}},\boldsymbol{n_{p_{ij}}})=-\sum\limits_{m=1}^{k} \omega{(\boldsymbol{p_{ij}},\boldsymbol{n_{m{p_{ij}}}})} f({d}_{\boldsymbol{n_{m{p_{ij}}}}})
\label{eq.esmoothness}
\end{equation}
where
\begin{equation}
\omega{(\boldsymbol{p_{ij}},\boldsymbol{n_{m{p_{ij}}}})}=\exp\left\{-\frac{||\mathcal{E}_m||_2^2}{{\sigma_d}^2}\right\}\exp\left\{-\frac{({d}_{\boldsymbol{n_{m{p_{ij}}}}}-{d}_{\boldsymbol{p_{ij}}})^2}{{\sigma_r}^2}\right\} 
\label{eq.weighting}
\end{equation}

\begin{figure}[!t]
	\centering
	\centering
	\subfigure[]
	{
		\includegraphics[width=0.225\textwidth]{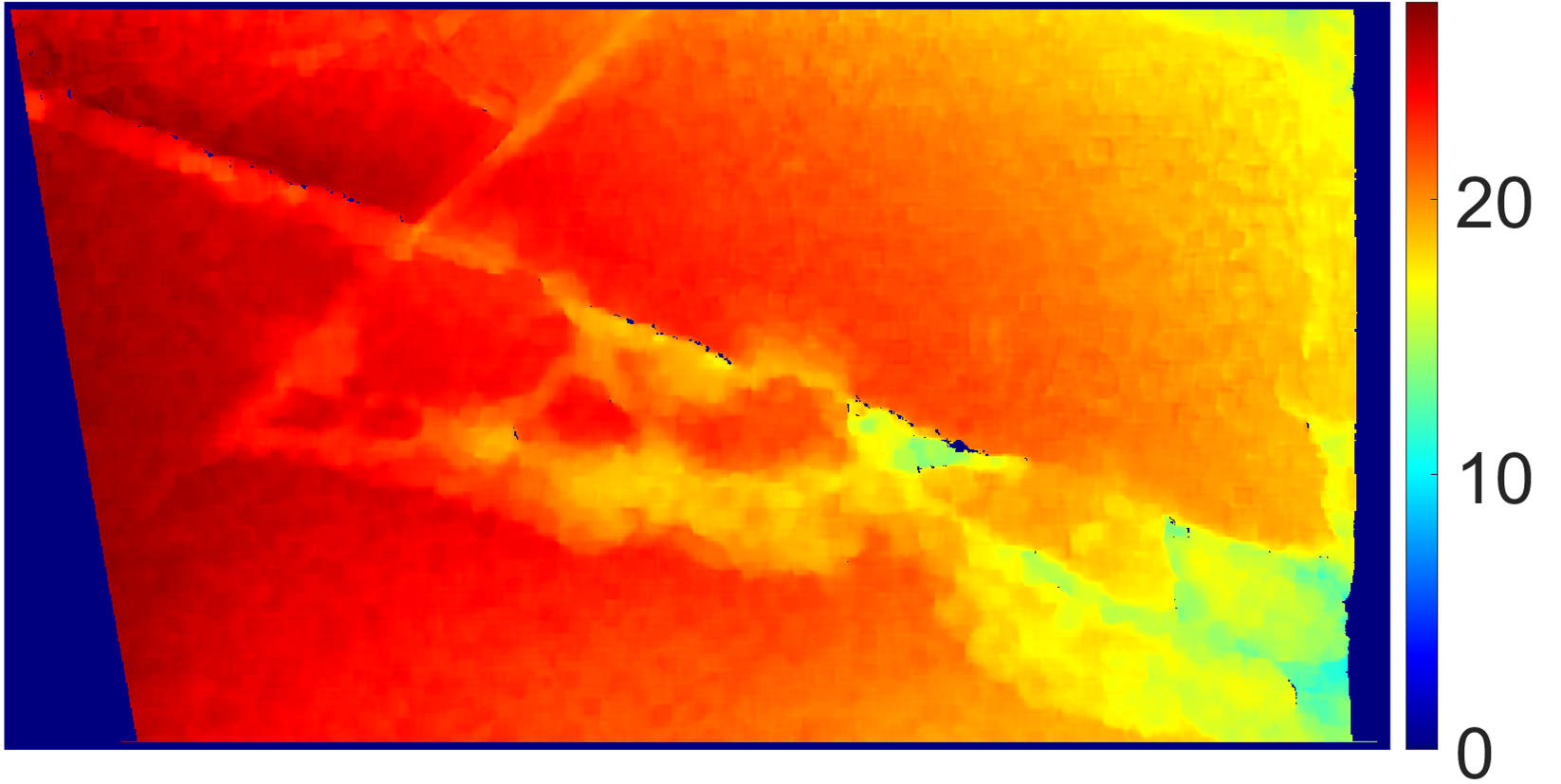}
	}
	\subfigure[]
	{
		\includegraphics[width=0.225\textwidth]{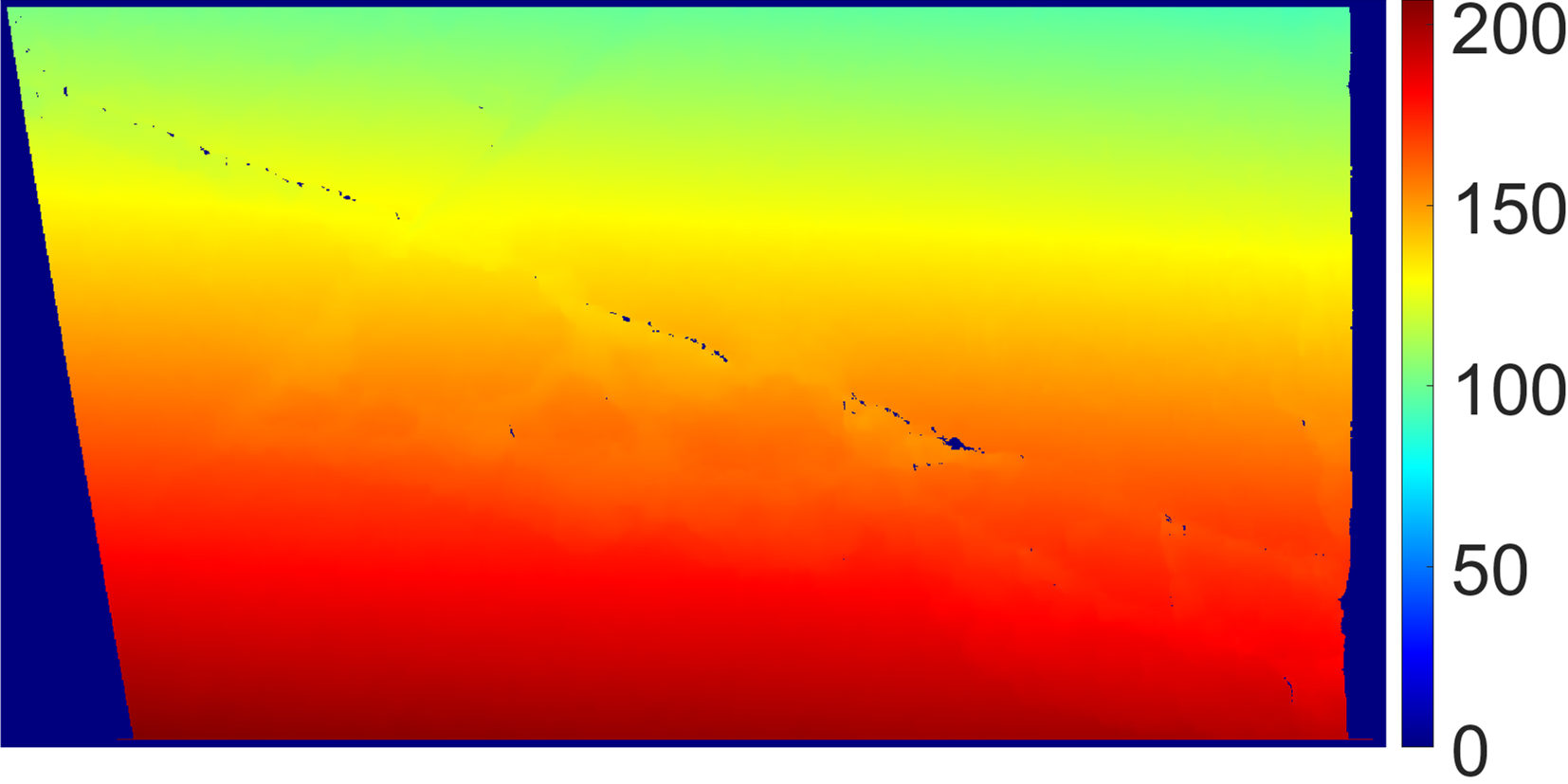}
	}
	\caption{Disparity map global refinement and post-processing. (a) subpixel disparity map after the third iteration. (b) post-processed disparity map. }
	\label{fig.disparity_map_gr}
		\vspace{-1.5em}
\end{figure}

The weighting coefficient $\omega$ is determined by both the spatial distance $||\mathcal{E}_m||_2$ between $\boldsymbol{n_{mp_{ij}}}$ and $\boldsymbol{p_{ij}}$ and the difference between ${d}_{\boldsymbol{n_{mp_{ij}}}}$ and ${d}_{\boldsymbol{p_{ij}}}$. $\sigma_d$ and $\sigma_r$ are two parameters used to control $\omega$ and they are respectively set to $1$ and $5$ in this paper. If ${d}_{\boldsymbol{n_{mp_{ij}}}}$ is similar to ${d}_{\boldsymbol{p_{ij}}}$, the weight for cost aggregation is higher. The energy function with respect to the correlation costs is updated iteratively. The subpixel disparity map is optimised by approximating the minima of the updated energy functions. In this paper, the proposed process is iterated three times, and the result after the third iteration is shown in Fig. \ref{fig.disparity_map_gr} (a).

\section{Post-Processing and 3D reconstruction}
\label{sec.post_processing_3d_rec}

\begin{figure}[b!]
	\centering
		\subfigure[]
	{
		\includegraphics[width=0.47\textwidth]{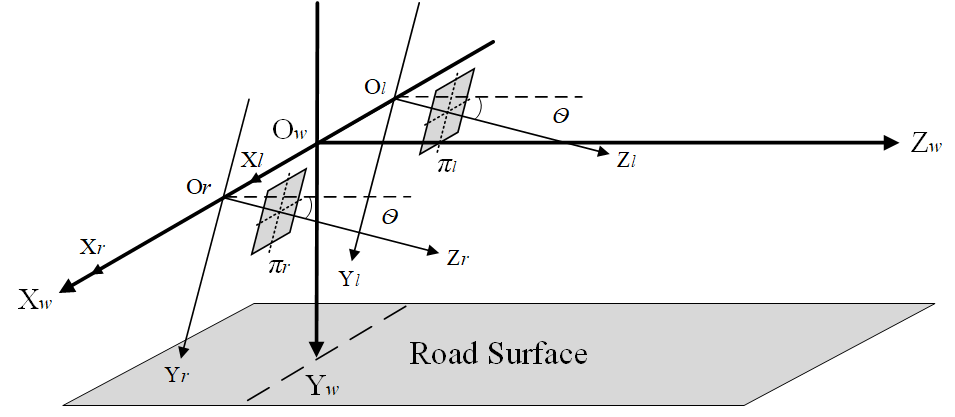}
	}
	\subfigure[]
	{
		\includegraphics[width=0.245\textwidth]{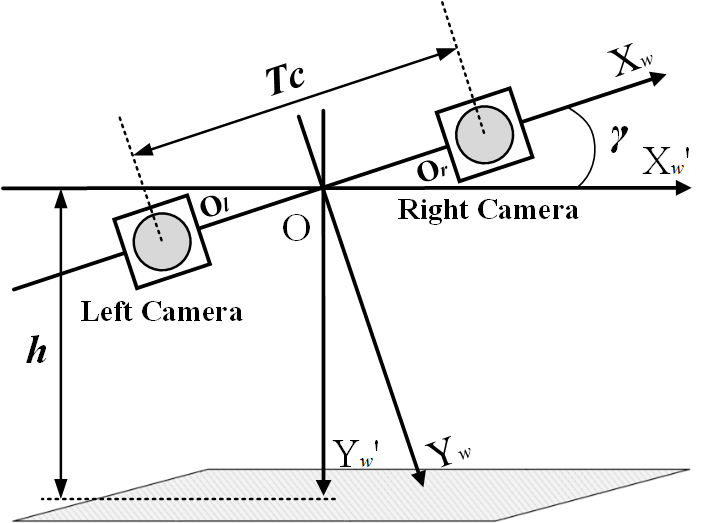}
	}
	\subfigure[]
	{
		\includegraphics[width=0.205\textwidth]{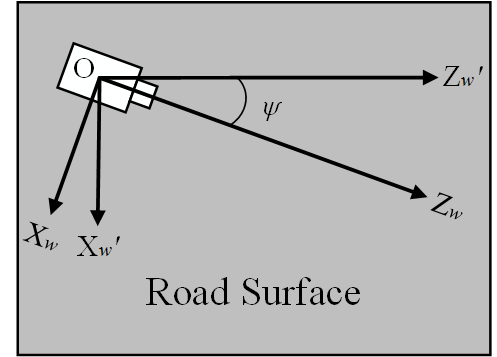}
	}
	
	\caption{Extrinsic rotations. (a) pitch angle $\theta$. (b) roll angle $\gamma$. (c) yaw angle $\psi$. $h$ is the height of the proposed binocular system. }
	\label{fig.extrinsic_rotations}
\end{figure}

Due to the fact that the perspective views have been transformed in section \ref{sec.perspective_transformation}, the estimated subpixel disparities on row $v$ should be added $\alpha_0+\alpha_1 v-\delta$ to obtain the post-processed disparity map which is illustrated in Fig. \ref{fig.disparity_map_gr} (b). Then, the intrinsic and extrinsic parameters of the stereo system are used to compute each 3D point $\boldsymbol{P_w}=[X_w,Y_w,Z_w]^\top$ from its projections $\boldsymbol{{p}_l}=[u_l,v_l]^\top$ and $\boldsymbol{{p}_r}=[u_r,v_r]^\top$, where $v_r$ is equivalent to $v_l$, and $u_r$ is associated with $u_l$ by disparity $d$.

For many state-of-the-art road model estimation algorithms, the effects caused by the non-zero roll angle (Fig. \ref{fig.extrinsic_rotations} (b)) are always ignored because the stereo cameras will not change significantly over time \cite{Ozgunalp}. However, the experimental set-up in this paper is installed manually and the roll angle may introduce a distortion on the v-disparity histogram. Therefore, the roll angle needs to be estimated for the initial frame to minimise its impact on the perspective transformation for the rest of the sequences. As in \cite{Ozgunalp}, the roll angle $\gamma$ can be estimated by fitting a linear plane ($d(u,v)=\gamma_0+\gamma_1u+\gamma_2v$) to a small patch from the near field in the disparity map and $\gamma=\arctan(-\gamma_1/\gamma_2)$. The pitch angle $\theta$ can be estimated by rearranging Eq. \ref{eq.relation_vd} as Eq. \ref{eq.pitch_angle}, where the parameters $[\alpha_0,\alpha_1]^\top$ have been approximated in section \ref{sec.perspective_transformation}.  The yaw angle $\psi$ shown in Fig. \ref{fig.extrinsic_rotations} (c) is assumed to be $0$. 

\begin{equation}
\theta=\arctan\left(\frac{1}{f}\left(\frac{\alpha_0}{\alpha_1}+v_0\right)\right)
\label{eq.pitch_angle}
\end{equation}

\begin{figure}[!b]
	\centering
	\includegraphics[width=0.32\textwidth]{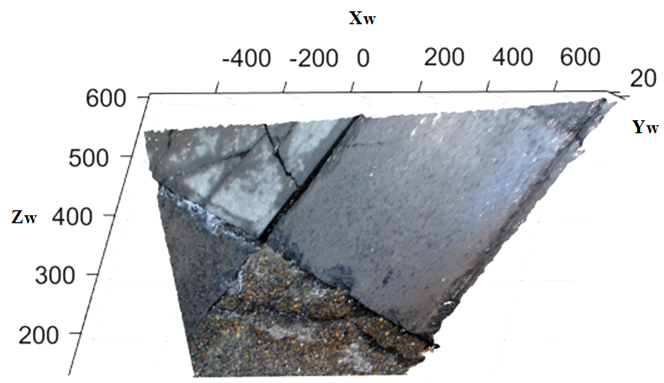}
	\caption{Road surface 3D reconstruction.}
	\label{fig.3d_recon_result}
\end{figure}

Each 3D point $[X_w, Y_w, Z_w]^\top$ can be transformed into $[X_w', Y_w', Z_w']^\top$ using Eq. \ref{eq.3d_transformation} \cite{Slabaugh1999}. The rotation matrix $\boldsymbol{R}=\boldsymbol{R_{\psi}}\boldsymbol{R_{\theta}}\boldsymbol{R_{\gamma}}$ is a SO(3) matrix. The rotation with $\boldsymbol{R}$ makes pothole detection much easier. The 3D reconstruction of Fig. \ref{fig.pt_results} (a) is illustrated in Fig. \ref{fig.3d_recon_result}.

\begin{equation}
\begin{bmatrix}
X_w'\\
Y_w' \\
Z_w'\\
\end{bmatrix}
=
\boldsymbol{R_{\psi}}\boldsymbol{R_{\theta}}\boldsymbol{R_{\gamma}}
\begin{bmatrix}
X_w\\
Y_w \\
Z_w\\
\end{bmatrix}
\label{eq.3d_transformation}
\end{equation}
where

\begin{equation}
\boldsymbol{R_{\psi}}
=
\begin{bmatrix}
\begin{array}{rrr}
\cos\psi & 0 & \sin\psi\\
0 & 1 & 0\\
-\sin\psi & 0 & \cos\psi\\
\end{array}
\end{bmatrix}
\end{equation}

\begin{equation}
\boldsymbol{R_{\theta}}
=
\begin{bmatrix}
\begin{array}{rrr}
1& 0 & 0\\
0 &\cos\theta &\sin\theta\\
0 & -\sin\theta & \cos\theta\\
\end{array}
\end{bmatrix}
\end{equation}

\begin{equation}
\boldsymbol{R_{\gamma}}
=
\begin{bmatrix}
\begin{array}{rrrr}
\cos\gamma & \sin\gamma & 0\\
-\sin\gamma & \cos\gamma & 0\\
0 & 0 & 1\\
\end{array}
\end{bmatrix}
\end{equation}

	\vspace{-0.5em}

\section{Experimental Results}
\label{sec.experimental_results}

In this section, we evaluate the performance of our proposed road surface 3D reconstruction algorithm both qualitatively and quantitatively. The algorithm is programmed in C language on an Intel Core i7-4720HQ CPU using a single thread. The following subsections detail the experimental set-up and the performance evaluation.

\subsection{Experimental Set-up}
\label{sec.experimental_setup}
\begin{figure}[!t]
	\centering
	
	\includegraphics[width=0.26\textwidth]{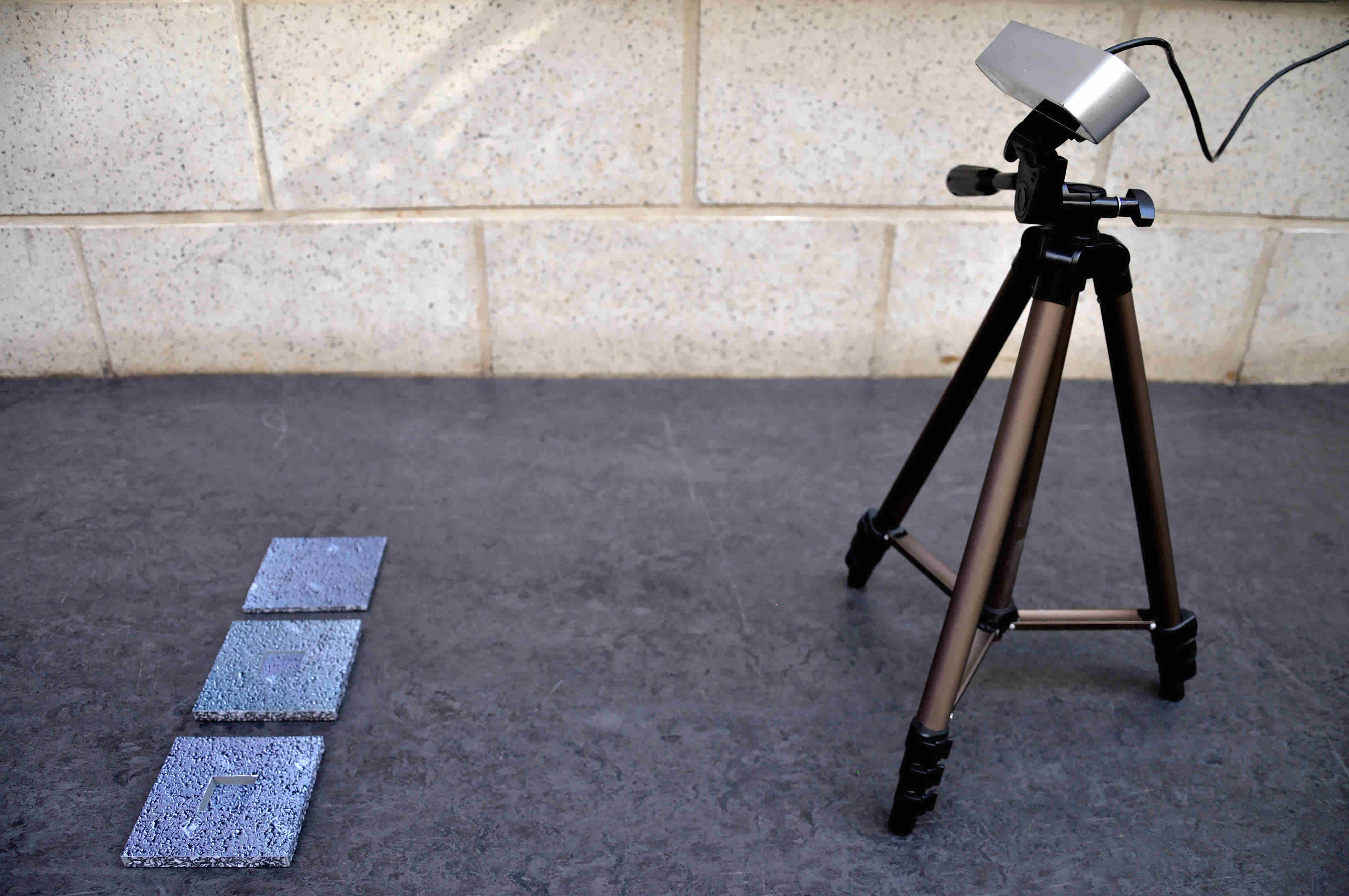}
	\caption{Experimental set-up.}
	\label{fig.exp_setup}
		\vspace{-1.3em}
\end{figure}

In our experiments, a state-of-the-art stereo camera from ZED Stereolabs is used to capture 1080p $(3840\times 1080)$ videos at 30 fps or 2.2K $(4416\times 1242)$ videos at 15 fps \cite{ZED}. The baseline is 120 mm. With its ultra sharp six element all-glass dual lenses and 16:9 native sensors, the video is $110^\circ$ wide-angle and able to cover the scene up to 20 m. An example of the experimental set-up is shown in Fig. \ref{fig.exp_setup}. The stereo camera is calibrated manually using the stereo calibration toolbox from MATLAB R2017a. The overall calibration mean error in pixels is 0.335.

\begin{figure}[!b]
	\centering
	
	\includegraphics[width=0.30\textwidth]{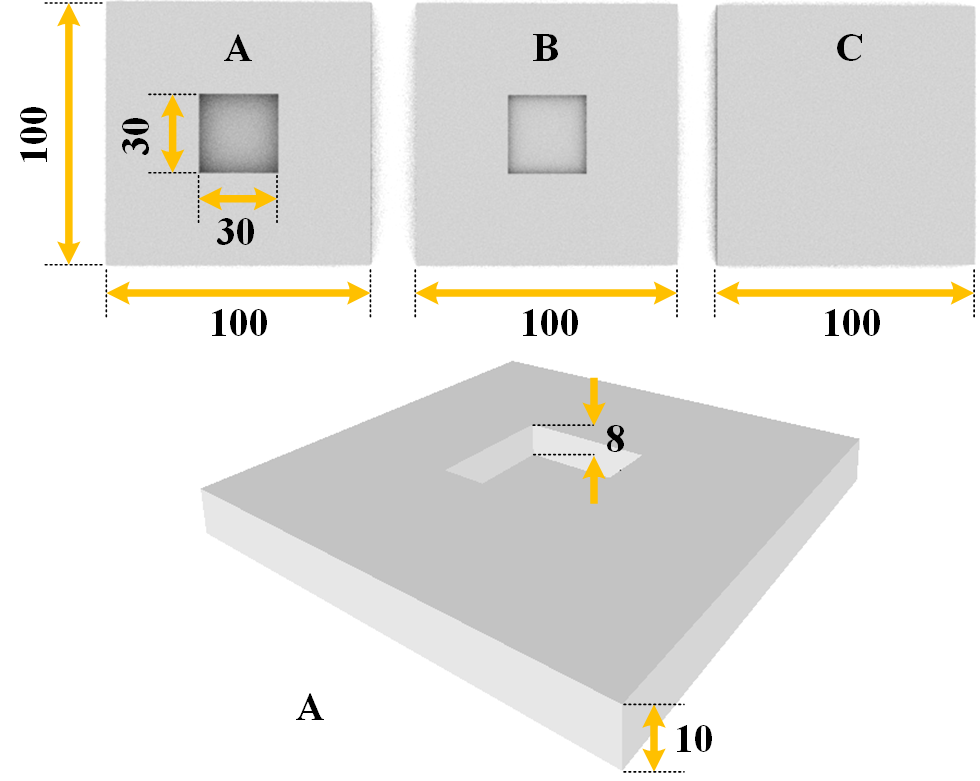}
	\caption{Designed 3D sample models. The unit is millimetre. }
	\label{fig.3d_sample_model}
\end{figure}
\begin{figure*}[!t]
	\centering
	\includegraphics[width=1\textwidth]{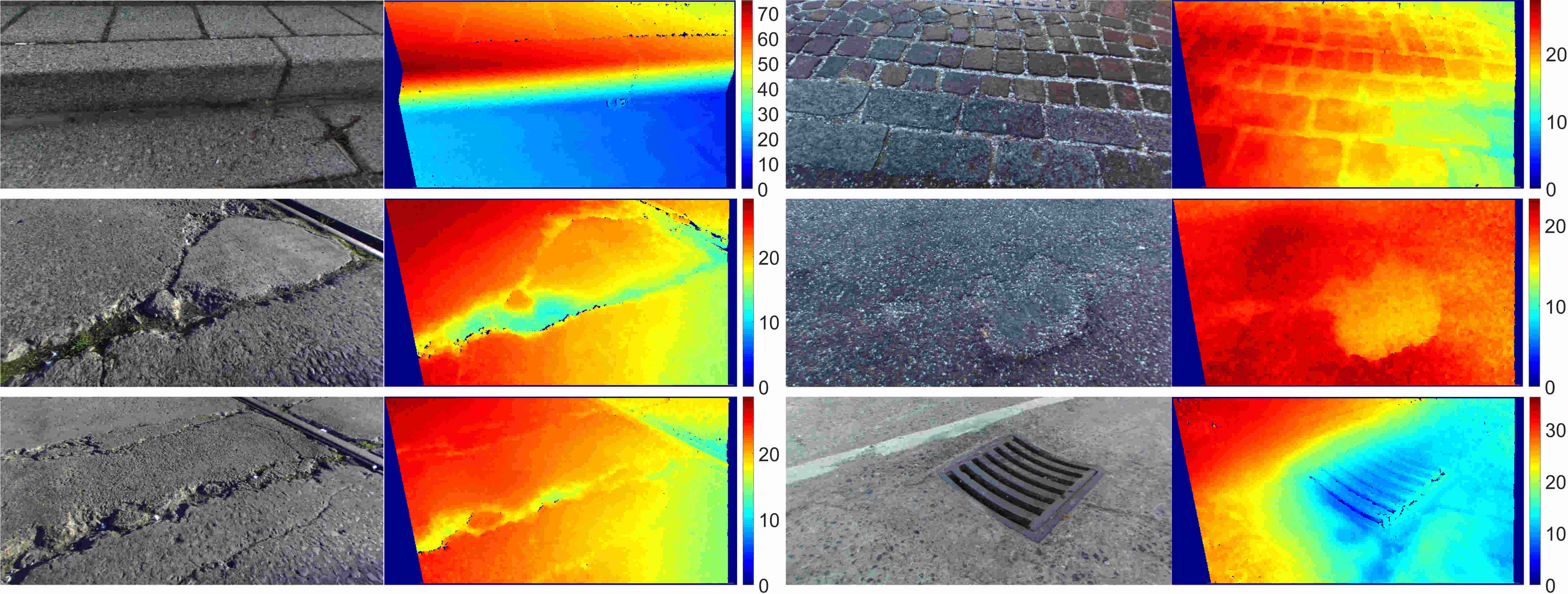}
	\caption{Experimental results. The first and third columns are the input left images. The second and fourth columns are the subpixel disparity map without post-processing.}
	\label{fig.road_surface_3d_exp_reslt}
\end{figure*}
\begin{table*}[!t]
	\begin{center}
		\vspace{0in}
		\footnotesize
				\caption{Design size and actual size of the sample models.}
		\label{table.3d_model_size}
		\begin{tabular}{|c|cc|cc|}
			\hline
			\multirow{2}{*}{Sample model} & \multicolumn{2}{c|}{Design size (mm$\times$mm$\times$mm)} & \multicolumn{2}{c|}{Actual size (mm$\times$mm$\times$mm)} \\
			\cline{2-5}
			& Model & Groove & Model & Groove \\
			\hline
			A & $100.00\times100.00\times10.00$ & $30.00\times30.00\times8.00$ & $99.97\times99.83\times10.31$ & $29.74\times30.01\times8.25$\\
			\hline
			B & $100.00\times100.00\times10.00$ & $30.00\times30.00\times3.00$ & $100.39\times100.10\times9.82$ & $30.28\times29.98\times3.52$\\
			\hline
			C & $100.00\times100.00\times5.00$ & n/a & $100.00\times99.98\times5.92$ & n/a\\
			\hline
		\end{tabular}
		\vspace{-2.5em}
	\end{center}
\end{table*}
To quantify the accuracy of the proposed algorithm, we designed three sample models A, B and C with different sizes. They are printed with a MakerBot Replicator 2 Desktop 3D Printer whose layer resolution is from 0.1 mm to 0.3 mm. Their top views and the stereogram of model A are illustrated in Fig. \ref{fig.3d_sample_model}, where A and B are designed with grooves to simulate potholes. To get the ground truth for our experiments, we measured the actual size of these models using an electronic vernier caliper. Both the design and actual sizes of the models are presented in Table \ref{table.3d_model_size}. Since the models are printed with a single colour, resulting in homogeneous areas, we attached them with a piece of paper with the texture of the road surface printed on it to avoid the ambiguities during stereo matching, as can be seen in Fig. \ref{fig.exp_setup}.

Using the above experimental set-up, we create three datasets (91 stereo image pairs) for the road surface 3D reconstruction.  Datasets 1 and 2 aim at road sceneries, and dataset 3 contains the sample models to help researchers qualify their reconstruction results. The datasets are available at: http://www.ruirangerfan.com.

The following subsections analyse the performance of our algorithm in terms of disparity accuracy, reconstruction accuracy and processing speed.

\subsection{Disparity Evaluation}
\label{sec.disaprity_evaluation}

\begin{figure}[!b]
	\centering
	
	\includegraphics[width=0.36\textwidth]{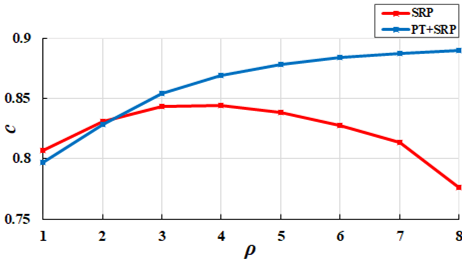}
	\caption{Comparison between SRP and PT+SRP in terms of the average of the highest correlation costs.  
	}
	\label{fig.pt_improvement}
\end{figure}
Some examples of the disparity maps are illustrated in Fig. \ref{fig.road_surface_3d_exp_reslt}.
Before estimating the disparity map, we transform the target image into its reference view, which greatly eliminates the perspective distortion for a GP between two images. Since the GP in the left and right images now looks similar to each other, the average of the highest correlation costs goes higher, which is depicted in Fig. \ref{fig.pt_improvement}. For stereo matching with only SRP, the average of the highest correlation increases gradually from $0.807$ ($\rho=1$) to $0.845$ ($\rho=4$). However, when $\rho$ goes above $4$, $c$ keeps decreasing. If we pre-process the input image pairs with the PT, the average of the highest correlation costs in the SRP stereo will grow gradually between $\rho=1$ and $\rho=8$. In this paper, our datasets are created with high-resolution images, and $\rho$ is proposed to be $5$. Compared with the conventional SRP stereo, the PT improves the average correlation cost with an increase of $0.05$. 

\begin{figure}[!b]
	\centering
	
	\includegraphics[width=0.49\textwidth]{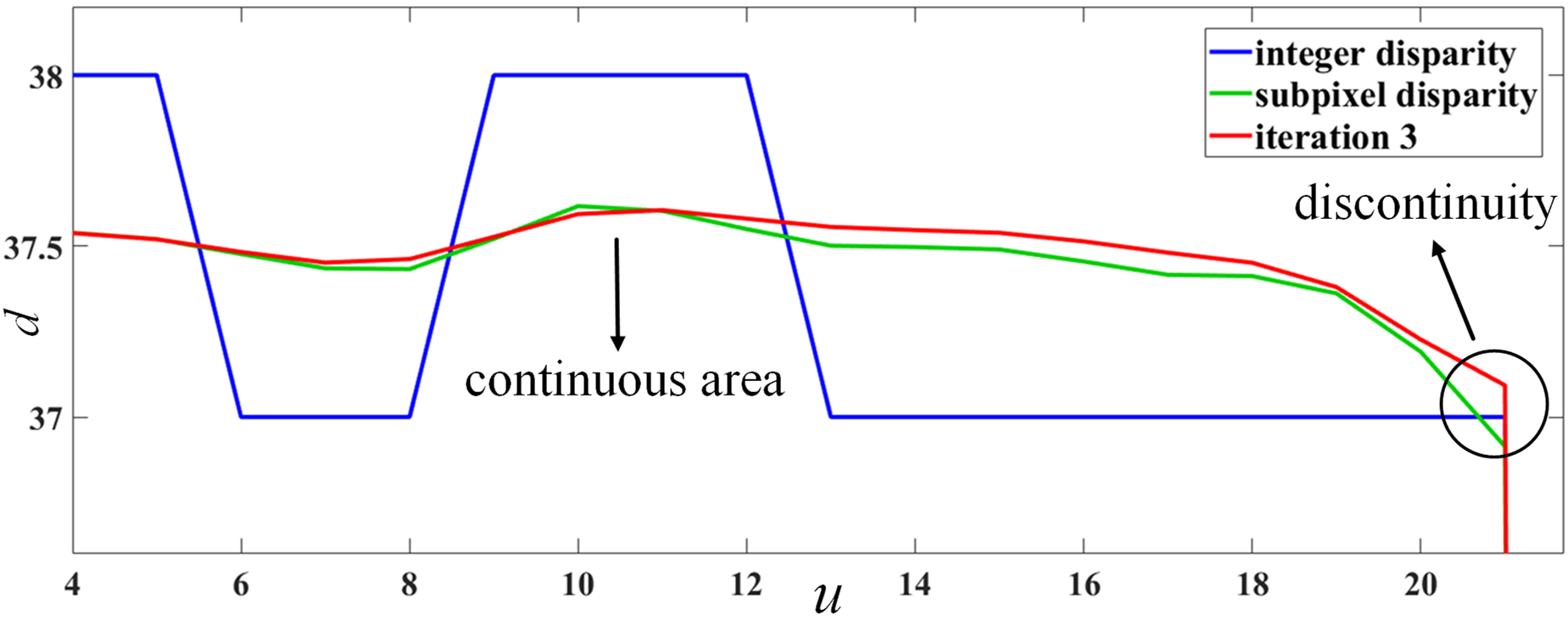}
	\caption{Evaluation of subpixel enhancement and disparity global refinement.  
	}
	\label{fig.gr_eva}
\end{figure}

\begin{figure*}[!t]
	\centering
	\includegraphics[width=1\textwidth]{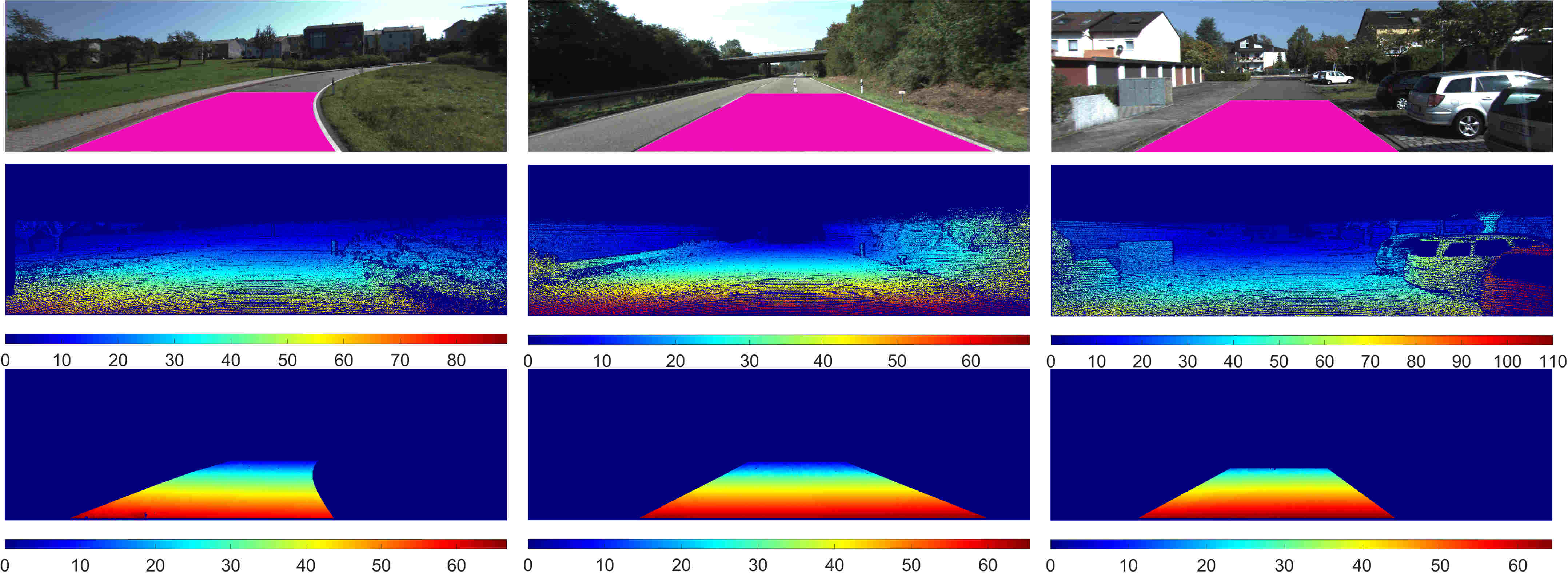}
	\caption{Experimental results of the KITTI stereo 2012 dataset. The first row shows the left images, where areas in magenta are our manually selected road surface. The second row shows the disparity ground truth. The third row shows the results obtained from the proposed algorithm.  
	}
	\label{fig.kitti_results}
\end{figure*}

\begin{table*}[!t]
	\begin{center}
		\vspace{0in}
		\footnotesize
				\caption{3D reconstruction measurement range.}
		\label{table.rec_experimental_results}
		\begin{tabular}{|c|ccccc|}
			\hline
			\multirow{2}{*}{Target} & \multicolumn{5}{|c|}{Measurement range (mm)} \\
			\cline{2-6}
			& $D\approx450 mm$ & $D\approx470 mm$ & $D\approx500 mm$ & $D\approx550mm$ & $D\approx650 mm$\\
			\cline{2-6}
			\hline
			Model A height & $09.72-10.21$ & $09.64-11.12$ & $10.31-12.19$ & $09.59-12.37$ & $08.99-12.62$\\
			Model B height & $09.86-10.32$ & $09.91-10.47$ & $10.07-11.25$ & $10.10-11.99$ & $10.86-12.36$\\
			Model C height & $04.62-05.54$ & $04.92-06.11$ & $05.72-06.93$ & $06.61-07.18$ & $06.69-07.54$\\
			Groove A depth & $07.77-08.44$ & $08.31-09.54$ & $05.92-09.17$ & $05.49-07.26$ & $09.37-11.83$\\
			Groove B depth & $02.21-05.12$ & $04.88-05.32$ & $04.97-06.51$ & $06.28-07.57$ & $05.29-06.63$\\
			\hline	
		\end{tabular}
		\vspace{-2.5em}
	\end{center}
\end{table*}
Furthermore, we select one row from the disparity map to evaluate the performance of subpixel enhancement and global refinement (see Fig. \ref{fig.gr_eva}). The integer disparity $d$ oscillates along the selected row and drops down abruptly when a discontinuity occurs. After the subpixel enhancement, the disparity $d$ is replaced with a better one $d_s$ between $d-1$ and $d+1$. The iterative global refinement further optimises the subpixel disparity map. After the third iteration, the disparities change more smoothly in a continuous area but interrupt suddenly when reaching a discontinuity.

Since the datasets we create only contain the ground truth of 3D reconstruction, the KITTI stereo 2012 dataset \cite{Andreas2012} is used to further evaluate the disparity accuracy of our algorithm. Some experimental results are illustrated in Fig. \ref{fig.kitti_results}. Due to the fact that the proposed algorithm only aims at reconstructing the road surface, we select a region of interest (see the magenta areas in the first row) from each image to evaluate the performance of our algorithm. The corresponding disparity results in the region of interest are shown in the third row. The percentage of error pixels (threshold: two pixels) is around $0.73\%$ and the average error in pixels is about $0.51$.   
\begin{figure}[b!]
	\centering
	\subfigure[]
	{
		\includegraphics[width=0.206\textwidth]{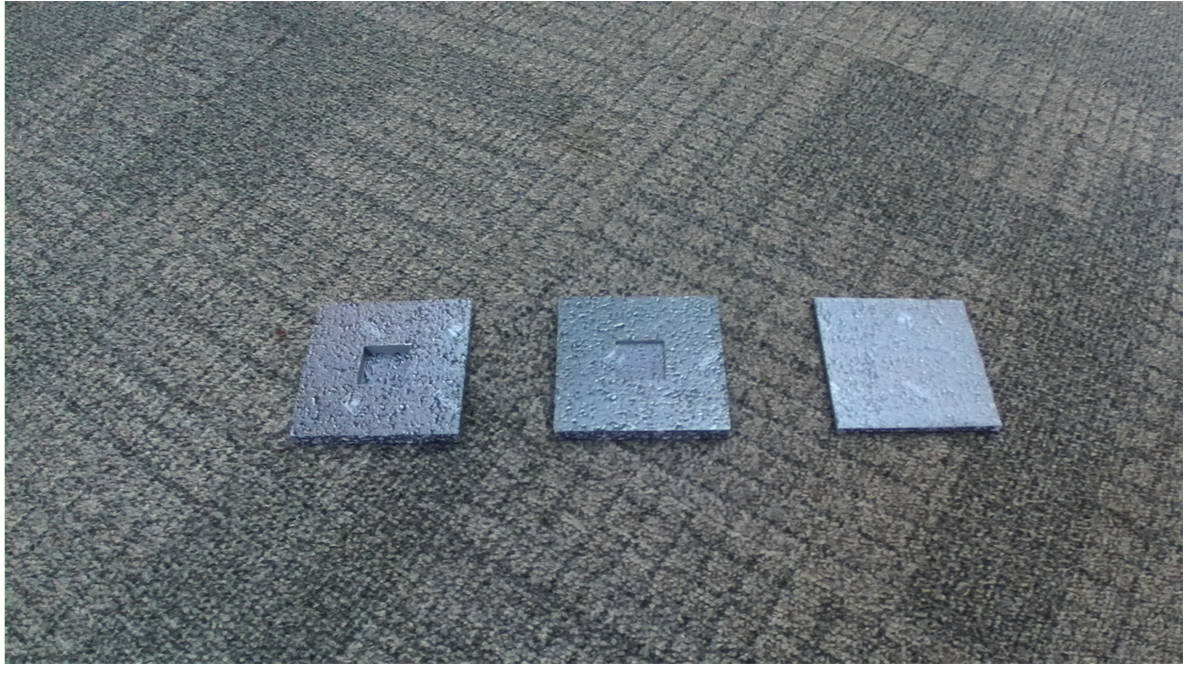}
	}
	\subfigure[]
	{
		\includegraphics[width=0.229\textwidth]{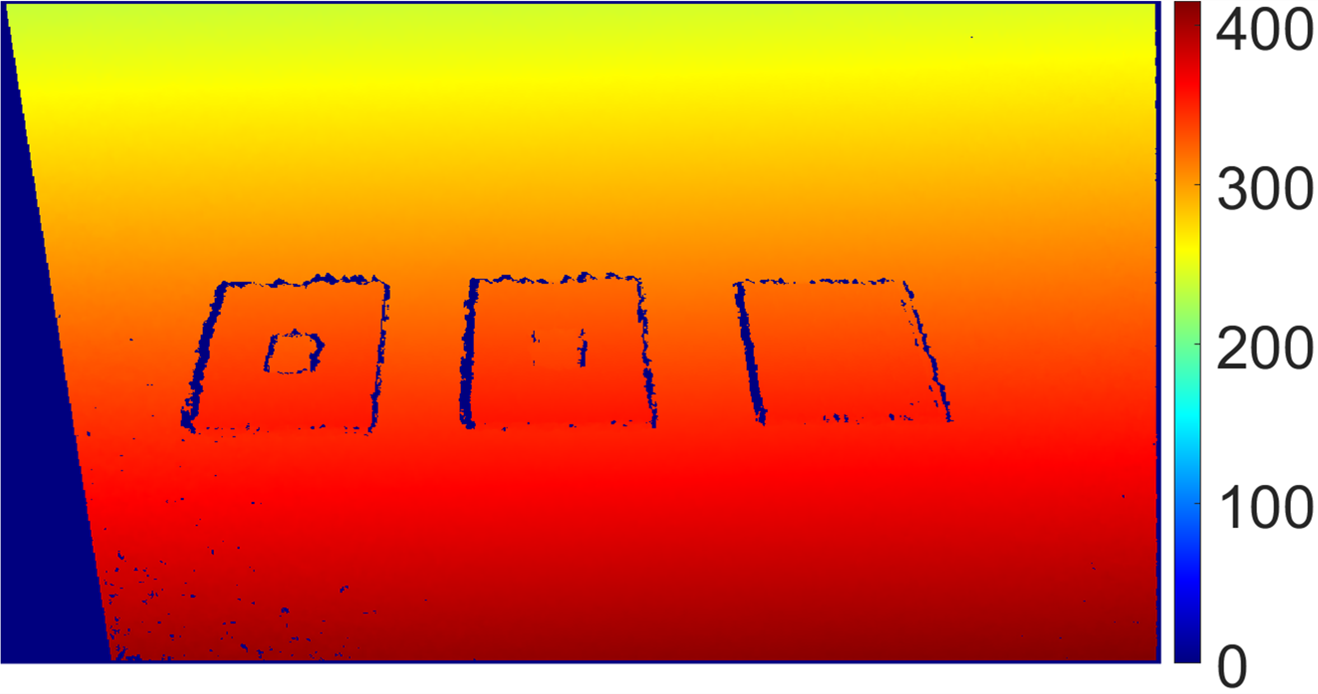}
	}
	\subfigure[]
	{
		\includegraphics[width=0.22\textwidth]{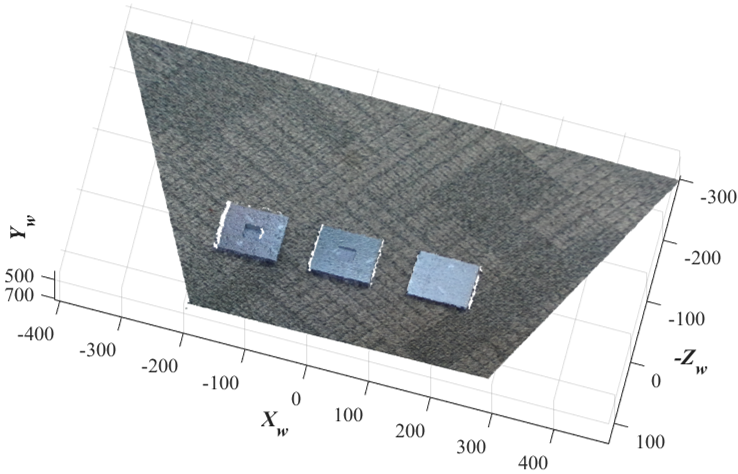}
	}
	\subfigure[]
	{
		\includegraphics[width=0.22\textwidth]{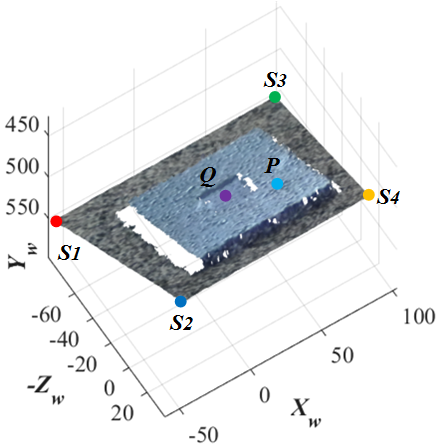}
	}
	\caption{Sample model 3D reconstruction. (a) left image. (b) subpixel disparity map with post-processing. (c) reconstructed scenery. (d) selected 3D point cloud which includes model B.}
	\label{fig.reconstruction_eva}
\end{figure}
\subsection{Reconstruction Evaluation}
\label{sec.reconstruction_evaluation}

To further evaluate the accuracy of the reconstruction results, we create dataset 3 (see section \ref{sec.experimental_setup} for details) with three different sample models. An example of the left image is illustrated in Fig. \ref{fig.reconstruction_eva} (a). The corresponding subpixel disparity map and 3D reconstruction are depicted in Fig. \ref{fig.reconstruction_eva} (b) and (c), respectively. We select a rectangular region which includes one of the sample models from Fig. \ref{fig.reconstruction_eva} (a), and the 3D reconstruction of this region can be seen in Fig. \ref{fig.reconstruction_eva} (d). A surface $\kappa_0 X_w + \kappa_1 Y_w + \kappa_2 Z_w + \kappa_3=0$ is fitted to four corners $S_1$, $S_2$, $S_3$ and $S_4$ of the selected region. Then, we select a set of random points $\boldsymbol{P_{1}}, \boldsymbol{P_{2}}, \dots, \boldsymbol{P_{n}}$ on the surface of the model and estimate the distances between them and the fitted road surface. These random distances provide the measurement range of the model height. Similarly, the groove depth can be estimated by computing the distances between a group of points $\boldsymbol{Q_{1}}, \boldsymbol{Q_{2}}, \dots, \boldsymbol{Q_{n}}$ in a groove and the model surface. Table \ref{table.rec_experimental_results} details the range of the measured model height and groove depth, where $D$ represents the approximated distance from the camera to sample models.

From Table \ref{table.rec_experimental_results}, the maximal absolute error of the 3D reconstruction is approximately 3 mm, and it increases slightly when $D$ increases. The reconstruction precision is inversely proportional to the depth \cite{Llorca2010}. Furthermore, since the baseline of the ZED camera is fixed and cannot be increased to further improve the precision, we mount it to a relatively low height and it is kept as perpendicular as possible to the road surface to reduce the average depth, which guarantees a high reconstruction accuracy.

\subsection{Processing Speed}
\label{sec.processing_speed}
\begin{figure}[!t]
	\centering
	
	\includegraphics[width=0.36\textwidth]{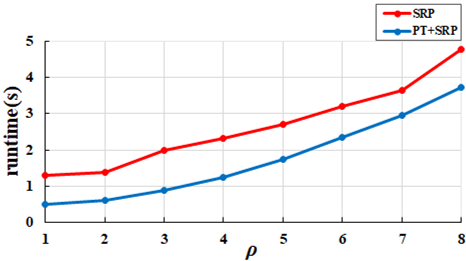}
	\caption{Comparison between SRP and PT+SRP in terms of the runtime.  
	}
	\label{fig.runtime_eva}
	\vspace{-1.6em}
\end{figure}
The algorithm is implemented in C language on an Intel Core i7-4720HQ CPU (2.6 GHz) using a single thread.
After the PT, each point on row $v$ in the target image is shifted $a_0+a_1v-\delta$ pixels to obtain a reference view, which greatly reduces the search range for stereo matching. The evaluation of the PT with respect to the runtime is illustrated in Fig. \ref{fig.runtime_eva}. The PT accelerates the processing speed of the SRP stereo when using different block sizes. When $\rho=5$, the processing speed is increased by over $36\%$. 
The runtime of different datasets is shown in Table \ref{table.rec_runtime}. Although the proposed algorithm does not run in real time, the authors believe that its speed can be increased in the future by exploiting the parallel computing architectures.

\begin{table}[!h]
	\begin{center}
		\vspace{0in}
		\footnotesize
				\caption{Algorithm Runtime.}
		\label{table.rec_runtime}
		\begin{tabular}{|c|c|c|c|}
			\hline
			Dataset & Frames & Resolution & Runtime (s)  \\
			\hline
			Dataset 1 & 35 & $1240\times609$ & 0.71\\
			\hline
			Dataset 2 & 35 & $1249\times620$  & 0.84\\
			\hline
			Dataset 3 & 21 & $2081\times1048$  & 2.23\\
			\hline
		\end{tabular}
		\vspace{-1.8em}
	\end{center}
\end{table}

\section{Conclusion and Future Work}
\label{sec.3d_conclusion}
 The main novelties of this paper include PT, CMV, and disparity map global refinement. We created three datasets and made them publicly available to contribute to 3D reconstruction-based pothole detection. The PT not only enhances the similarity of a GP between two images but also reduces the search range for stereo matching. This helps the SRP stereo perform more accurately and efficiently. The CMV further offsets the insufficient propagation in the SRP stereo and guarantees the feasibility of parabola interpolation in the subpixel enhancement phase. By iteratively minimising the energy with respect to the interpolated parabolas, the subpixel disparity map is optimised. The disparities in a continuous area become more smooth, but they are preserved when discontinuities occur. The maximal absolute error of the 3D reconstruction is around 3 mm, which satisfies the requirement of millimetre accuracy for on-road damage detection. Furthermore, due to the high precision of the proposed system, users can apply it to road surface SLAM (Simultaneous Localisation and Mapping) for many smart city applications.

However, the propagation strategy in the proposed algorithm makes it difficult to fully exploit the parallel computing architecture of the graphics cards to estimate disparity maps. Therefore, we aim to come up with a more efficient SRP strategy which can be adapted for different platforms. Furthermore, errors in stereo calibration always affect the precision of the stereo matching dramatically. Hence, we aim to design a self-calibration algorithm to enhance the robustness of our proposed stereo vision system, and the reconstructed sceneries will be used for 3D pothole detection.

\ifCLASSOPTIONcaptionsoff
  \newpage
\fi
\label{sec:refs}
\bibliographystyle{IEEEbib}

\end{document}